\documentclass[10pt]{article} 
\usepackage{tmlr-style-file/tmlr}

\usepackage{amsmath,amsfonts,bm}

\def\eqref#1{equation~\ref{#1}}

\def\1{\bm{1}}

\DeclareMathAlphabet{\mathsfit}{\encodingdefault}{\sfdefault}{m}{sl}
\SetMathAlphabet{\mathsfit}{bold}{\encodingdefault}{\sfdefault}{bx}{n}

\usepackage{hyperref}
\usepackage{url}
\usepackage{tabularx}
\usepackage{xparse}
\usepackage{seqsplit}
\usepackage{amsmath}
\usepackage{graphicx}
\usepackage{float}
\usepackage{booktabs}
\usepackage{ragged2e}
\usepackage{longtable}
\usepackage{pgfplots}
\usepackage{tabularx}
\usepackage{array}
\usepackage{subcaption}
\usepackage{multirow}
\usepackage{amssymb}
\usepackage{adjustbox}
\usepackage{makecell}
\usepackage{placeins}
\usepackage{CJKutf8}
\usepackage{longtable, booktabs, xcolor}

\title{Building a Taiwanese Mandarin Spoken Language Model: \\ A First Attempt}

\author{\name  Chih-Kai Yang*,
\name  Yu-Kuan Fu*,
\name Chen-An Li*,
\name Yi-Cheng Lin*,
\name Yu-Xiang Lin*,
\name Wei-Chih Chen*,
\name Ho Lam Chung*,
\name Chun-Yi Kuan*,
\name Wei-Ping Huang*,
\name Ke-Han Lu*,
\name Tzu-Quan Lin*,
\name Hsiu-Hsuan Wang*,
\name En-Pei Hu*,
\name Chan-Jan Hsu$^\dagger$,
\name Liang-Hsuan Tseng$^\dagger$,
\name I-Hsiang Chiu,
\name Ulin Sanga,
\name Xuanjun Chen,
\name Po-chun Hsu,
\name Shu-wen Yang,
\name Hung-yi Lee
\AND
* indicates an equally contributing first author.
\AND
$^\dagger$ indicates an equally contributing second author.
 }

\begin{document}

\maketitle

\begin{abstract}
This technical report presents our initial attempt to build a spoken large language model (LLM) for Taiwanese Mandarin, specifically tailored to enable real-time, speech-to-speech interaction in multi-turn conversations. Our end-to-end model incorporates a decoder-only transformer architecture and aims to achieve seamless interaction while preserving the conversational flow, including full-duplex capabilities allowing simultaneous speaking and listening. 
The paper also details the training process, including data preparation with synthesized dialogues and adjustments for real-time interaction.
We also developed a platform to evaluate conversational fluency and response coherence in multi-turn dialogues.
We hope the release of the report can contribute to the future development of spoken LLMs in Taiwanese Mandarin.
\end{abstract}

\section{Introduction} 
This is a technical report on the practice of training large language models (LLMs) for spoken Taiwanese Mandarin. 
The report serves as a record of the work completed to date.
As a work-in-progress paper, the current version is not yet ready for conference submission. However, we believe that some ideas and findings from this project may inspire future work. Therefore, we have made the report publicly available. If you find the ideas or findings in this report useful for your project, please cite this technical report.

\subsection{Goal of the Project}
The project aims to train a spoken LLM that can engage in real-time speech-to-speech conversations. The model takes a system prompt as input and, based on that prompt, facilitates seamless spoken interactions. Ultimately, the aim is to create an experience where conversing with the AI feels as natural as talking to a real person on Google Meet.

In terms of interaction, there are several goals we aim to achieve:
\begin{itemize}
    \item Speech-to-speech, multi-turn conversations
    \item Mandarin conversation with a Taiwanese accent
    \item The model should be capable of seamless, full-duplex communication with humans. This means the model can process and respond in real time, allowing for interruptions and simultaneous speaking and listening. 
\end{itemize}
In the past, spoken dialogue has been achieved through a cascade framework, which combines automatic speech recognition (ASR), LLM, and text-to-speech (TTS) to facilitate speech-to-speech interaction. Here, we aim to build an end-to-end model. Since the model is end-to-end, we hope it can accomplish several tasks that a cascade model cannot. For example, it should be able to understand input audio beyond just the text, including prosody and speaker characteristics, and respond appropriately~\citep{lin-etal-2024-advancing}. It should also recognize non-verbal vocalizations (e.g., laughter, crying) and respond with the corresponding non-verbal sounds. Compared to a cascade model, an end-to-end model has the potential to interpret environmental sounds. However, we did not implement or thoroughly evaluate all of these capabilities due to time constraints, so we are unsure how well the model currently performs in these areas. Nonetheless, the end-to-end model has this potential, and we can extend the current work to incorporate these abilities in the future.

Although GPT-4's voice mode can achieve seamless interaction with speech, it is not open-sourced, so we cannot know how it is achieved. We hope this technical report can shed light on the development of spoken LLMs in the future, helping to accelerate the creation of open-source spoken LLMs for the community.
Our model will soon be open-sourced.
The open-source model most similar to our goal is Moshi~\citep{défossez2024moshispeechtextfoundationmodel}, but it primarily supports English and cannot understand or generate Mandarin.

\subsection{Technical Contributions in This Project}
We believe this technical report may contain some technological contributions, though we have not fully verified them. Below, we briefly mention the potential contributions.

For the model side, we demonstrate the potential to achieve real-time full-duplex machine-human speech communication using a typical decoder-only transformer architecture in Section~\ref{sec:network}. 
Existing speech-language models for generation, such as LLaMA-Omni~\citep{fang2024llamaomniseamlessspeechinteraction}, Mini-Omni~\citep{xie2024miniomnilanguagemodelshear}, and Moshi~\citep{défossez2024moshispeechtextfoundationmodel}, require modifications to the standard transformer model. In contrast, we generate audio based on the standard decoder-only transformer architecture. 
Traditionally, full-duplex communication is achieved using a dual-channel system -- one channel for listening and one for speaking~\citep{nguyen-etal-2023-generative,ma2024languagemodellistenspeaking}. However, we show that even a typical transformer architecture can achieve full-duplex communication. 
Since our model uses the same architecture as text LLM, we can directly use a text-based LLM for initialization and fine-tune it from there. Ideally, the speech model will inherit the text model's capabilities. 
Since it is possible to collect much more text data than speech data, initializing the speech model from a text model and leveraging the capabilities of a text LLM is critical to the success of a spoken LLM.
Additionally, because we do not modify the network architecture of the text LLM, the model can benefit from the wide range of frameworks and toolkits designed to enhance text LLMs. This also means it can be easily deployed and accelerated.

The data preparation and model training pipeline will be described in Section~\ref{sec:data} and \ref{sec:training}.
We originally intended to use a large amount of dialogue data collected from the Internet. However, we found that using real dialogues from the Internet negatively impacted the model’s performance. Therefore, we opted to use synthesized data instead. First, we used a text-based LLM to generate dialogues, then applied a TTS model to vocalize the generated dialogues. 
To ensure that the TTS output sounded natural with a \textbf{Taiwanese accent}, we used a TTS model developed by another project\footnote{The technical report of the project will be available soon}.
Our goal is for the model to generate realistic dialogue that aligns with the emotional content of the conversation. For example, when a character says, “I feel sad,” the speech should sound appropriately sad. Typical TTS systems, especially commercial APIs, often struggle to capture such nuances, so we rely on our own TTS models. To create more realistic dialogues, we even generate dialogues with interruptions, which will be further explained in the following discussion. 

Even with a well-trained model, numerous implementation challenges remain in achieving real-time conversation. All the difficulties, solutions, and tips will be further discussed in  Section~\ref{sec:system}.

In Section~\ref{sec:evaluation}, we will discuss how to evaluate spoken LLMs.
Currently, there is no standard method for evaluating spoken LLMs. Commonly used benchmarks, such as Dynamic-SUPERB~\citep{DynamicSUPERBICASSP23, dynamic-superbv2}, are instruction-following-based evaluations. 
However, we also want to assess how well the model can engage in communication, which no existing benchmark adequately measures. 
To address this, we developed a platform to facilitate communication between two chatbots. We also created an interface for users to interact with the bot, which simulates a smooth, natural conversation.

\section{Network Architecture} \label{sec:network}

\subsection{Overview}

\begin{figure}
    \centering
    \includegraphics[width=0.825\linewidth]{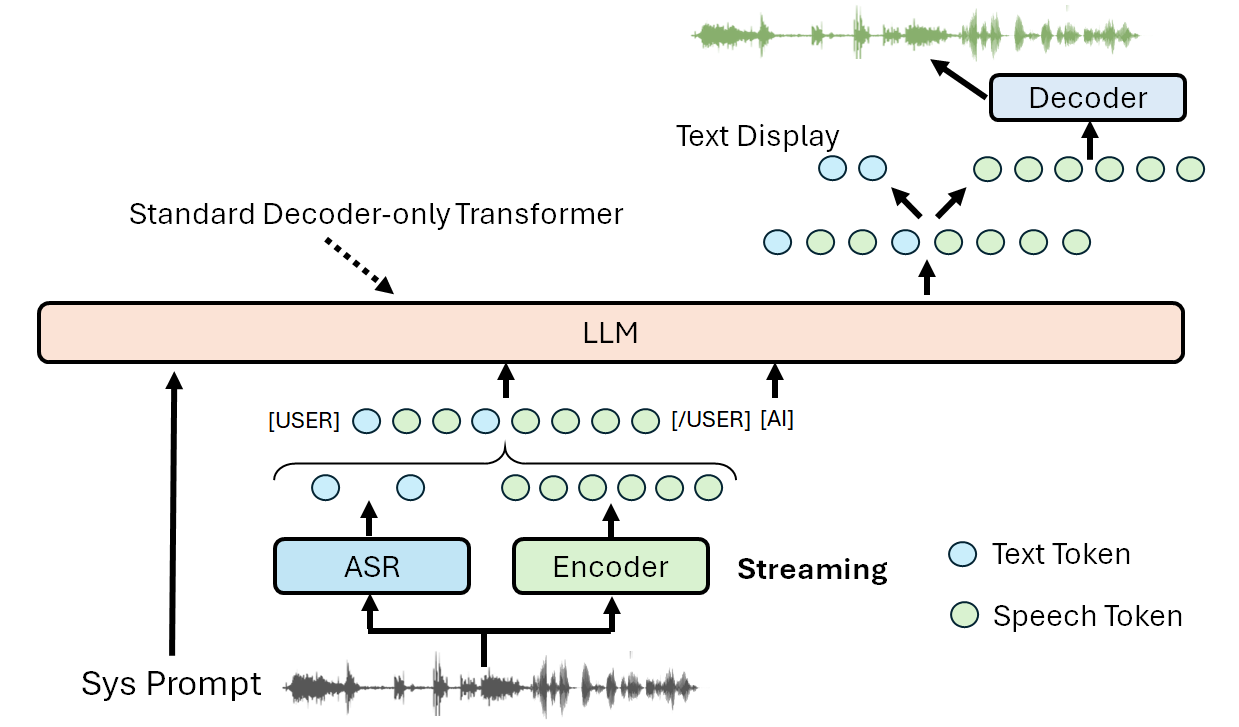}
    \caption{An overview of this project's spoken LLM network architecture.}
    \label{fig:framework}
\end{figure}

There has already been some research on spoken LLMs, with models publicly available~\citep{défossez2024moshispeechtextfoundationmodel,fang2024llamaomniseamlessspeechinteraction,xie2024miniomnilanguagemodelshear}\footnote{To be more specific, only Moshi~\citep{défossez2024moshispeechtextfoundationmodel} can enable seamless interaction. The others do not have the same capabilities.}.
However, these models are based on different network designs, and there is still no standard architecture for spoken LLMs. There remains significant research space for exploring various possibilities regarding the optimal architecture for a spoken LLM. 
The architecture of our model is shown in Figure~\ref{fig:framework}, which differs from previous work.
We will give an overview in this subsection and provide more details in the following subsections.

The main body of the model is a decoder-only transformer, with the same architecture as a text LLM, and it is initialized by a text LLM. The transformer takes a text system prompt as input and can also process speech as input and output speech. It supports multiple conversational turns. The model's architecture is identical to typical text models, which reduces additional development costs and helps the text LLM retain its original capabilities when fine-tuned for speech.
More details are in Section~\ref{subsec:llm_backbone}.

The input to the transformer uses a combination of ASR and speech encoder outputs. The ASR provides text tokens, while the speech encoder produces speech units. In the diagram, blue circles represent text tokens, and green circles represent speech units. 
The process for handling the speech signal as input to the standard transformer will be discussed in Section~\ref{subsec:encoder}. 

The transformer's output consists of a combination of text and speech units. The text is displayed on the screen, while the speech units are synthesized into speech through a speech decoder.
Section~\ref{subsec:decoder} will provide more details.

One of the most intriguing aspects of the model in Figure~\ref{fig:framework} is its ability to achieve seamless interaction.
For text-based interaction, a user types ``enter'' to indicate the end of the input. 
However, this method cannot be used for seamless speech dialogue. On the other hand, simply waiting for a long enough silence before responding can result in an unnatural and delayed conversation. 
How does the model know when the input token sequence has ended and it needs to start generating a response (that is, start talking)?
To our surprise, the standard text LLM architecture can enable turn-taking in conversations without much additional effort. 
This wasn't intuitive at first glance, which will be further described in Secton~\ref{subsec:duplex}.


\subsection{LLM Backbone}
\label{subsec:llm_backbone}
We utilized LLaMA-3.1 8B~\citep{llama3} as the LLM backbone of our spoken LLM and injected the speech capability into the backbone with speech-related training, which will be detailed in Section \ref{sec:training}.

In addition to having the ability to engage in speech-based conversations, we believe that it is essential for the users to be allowed to make the model response only in text or in speech as well, depending on their needs. It is also crucial for the model to have the ability to take on different roles and solve a wide variety of tasks. Without additional modification of the architecture of the backbone model, we surprisingly found that such controls could be achieved with proper design of system prompt format. 

We begin by clarifying the ``modalities'' referred to in our model. It is important to note that since our LLM backbone does not directly generate waveform or mel-spectrograms, the model's output is entirely tokens. However, to enable the generation of speech responses, special tokens representing discrete speech units have been added to the model's vocabulary, which a decoder then uses to synthesize speech. Therefore, the term ``modalities'' in this context does not refer to the actual output modalities of the LLM, which are purely tokens. Instead, it refers to the different combinations of the original text tokens and the newly introduced special tokens representing speech units. We maintain this terminology throughout the paper.

There are three kinds of possible modalities for the input and output of our LLM backbone: (1) \textbf{text}: pure text tokens only; (2) \textbf{unit}: only the speech unit sequence; (3) \textbf{hybrid}: the interleaved sequences of the text tokens and the corresponding speech units. We use the system prompts to provide information on the input/output modalities of the model.

Specifically, we incorporated a prefix with a format like ``Modality: \{User: text, Machine: speech\}'' in the system prompts of training data, where ``User'' and ``Machine'' identified the modalities of the input and the model's response, respectively\footnote{In actual implementation, we used the term ``speech'' for the ``hybrid'' modality in the system prompts.}. 
Using the aforementioned example, the previously shown prefix indicated that the input from the user and the model response would be in text and speech (interleaved sequences of text tokens and speech units), respectively. We prepared the training data and ensured the modality of the input and the label of each data sample followed the specification of the corresponding prefix in the system prompt. With this methodology, our preliminary experiments showed that the model could precisely generate the output in the modality specified by the user in the system prompts during inference, demonstrating strong proficiency in modality control and enhancing the convenience and user experience.

The other crucial part of the system prompts is role control, i.e., task control. As a versatile model, an ideal spoken LLM should be able to take on various roles and solve a wide range of tasks for users, similar to an ordinary text-based LLM. Therefore, we incorporated this part into the model's system prompts, in addition to the modality control prefix. A complete system prompt is the concatenation of the modality control prefix and the role-playing control instruction, resembling ``Modality: \{User: text, Machine: speech\} You are a ChatBot. Have a fun chat with the user.'' In this example, the modality control prefix indicates that input and output modalities are text and hybrid sequences, and the role-playing control instruction specifies the role of the model as a ChatBot whose task is having an interesting conversation with the user. A highly diverse set of role-playing instructions in Taiwanese Mandarin was employed in the training data to make the role controls more precise.

\subsection{Streaming ASR and Encoder}
\label{subsec:encoder}

As we aim to develop a spoken LLM that supports seamless conversation, the system should support real-time interaction. Therefore, both ASR and encoder in Figure~\ref{fig:framework} have to be streaming models, ensuring the input of the LLM backbone can be processed in real time. 

Our streaming ASR model here is a streaming version of Whisper. Technical details will be explained in Section~\ref{sec:streaming_ASR}.
For the speech encoder, to achieve shorter processing times, we apply a sliding window to the input audio signal. We use a non-causal, self-supervised speech encoder, with a carefully chosen short window size to ensure the model can still generate meaningful embeddings for quantization. 

Whenever the ASR model generates a text token or the speech encoder produces a speech unit, it is sent to the transformer, which takes a hybrid of these text tokens and speech units as input. The interleaving of text and speech units is based on the timing of the ASR and speech encoder outputs.
During training, we aim to train the model to function with a hybrid of text and speech units, as well as with speech units alone.
We still include ASR in the complete framework to enhance its overall robustness because state-of-the-art ASR systems typically have access to much more significant amounts of speech data than what is available for our spoken LLM. However, it should be noted that our model can still obtain satisfactory performance without relying on the ASR module here, as shown in Section~\ref{subsec:results}.

\subsection{Decoder} 
\label{subsec:decoder} 
To enable the generation of both text and speech, the LLM backbone processes an interleaved sequence of text and speech units, producing a combined output. From this output, we extract the speech units, which are subsequently passed through a speech decoder to synthesize the speech signal.

Specifically, we leverage a denoising diffusion framework, which iteratively refines the noise signal, transforming it into a mel-spectrogram over multiple diffusion time steps. Inspired by~\citet{wang2023neural}, we incorporate a reference speaker's sound prompt during training. This sound prompt serves as a conditioning factor to control the timbre of the synthesized speech, allowing for higher speaker control abilities. 

The framework starts from two convolution encoders, one for speech units and the other for handling reference audio prompts. The features generated by these two encoders are then concatenated and fed into a series of Wavenet-like~\citep{van2016wavenet} blocks to reconstruct the mel-spectrogram. Once the mel-spectrogram has been produced, we feed it into HiFi-GAN~\citep{NEURIPS2020_c5d73680}, which converts the mel-spectrogram back into a high-fidelity speech waveform, completing the speech synthesis process.

\begin{figure}[t]
    \centering
    \begin{subfigure}[b]{\linewidth}
        \centering
        \includegraphics[width=\linewidth]{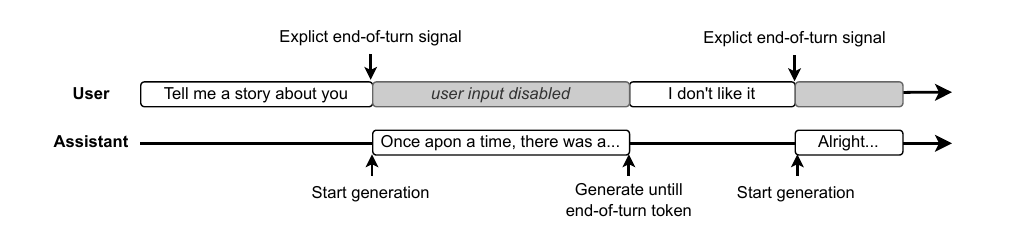}
        \caption{Conventional turn-based interaction relies on explicit end-of-turn signals (e.g., pressing enter, clicking send button) to manage turn-taking. User input is disabled during model generation, creating a rigid back-and-forth pattern.}
        \label{fig:duplex-1}
    \end{subfigure}
    
    \begin{subfigure}[b]{\linewidth}
        \centering
        \includegraphics[width=\linewidth]{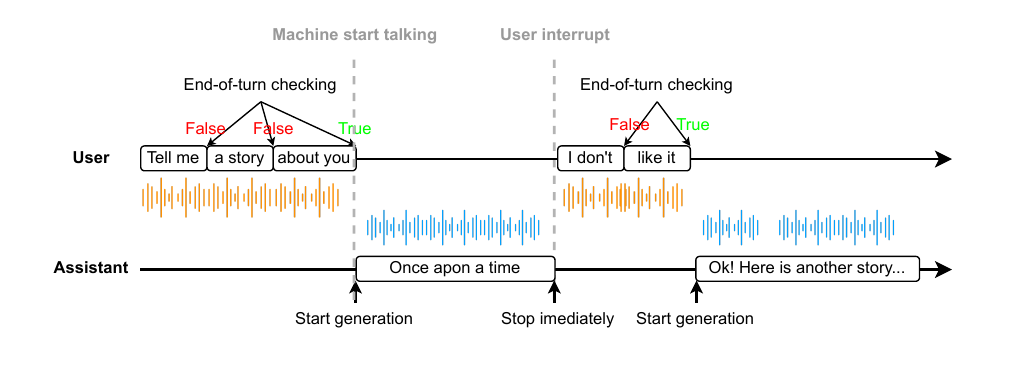}
        \caption{Our duplex interaction system enables fluid conversation by continuously checking for end-of-turn signals and user interruptions, allowing for natural turn-taking.}
        \label{fig:duplex-2}
    \end{subfigure}

    \begin{subfigure}[b]{\linewidth}
        \centering
        \includegraphics[width=\linewidth]{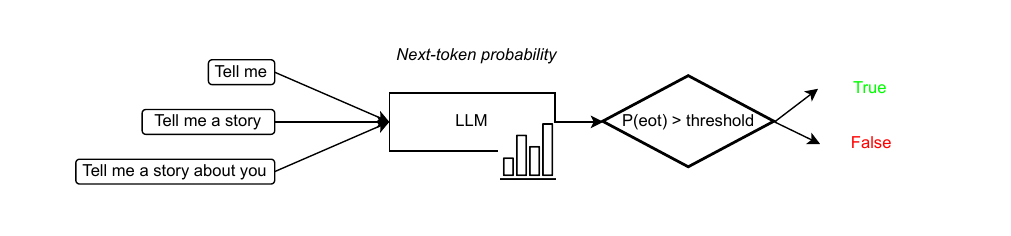}
        \caption{The end-of-turn checking process utilizes the language model's next-token prediction capability. For each accumulated input chunk, the system evaluates the probability of an end-of-turn token to determine if the user has completed their turn.}
        \label{fig:duplex-3}
    \end{subfigure}
    \caption{Overview of different interaction patterns and our proposed end-of-turn detection method.}
    \label{fig:overview}
\end{figure}
\subsection{Duplex Interaction} \label{subsec:duplex} 

The ability of language models to engage in seamless, real-time communication has been a significant challenge in developing user-friendly voice agents. Typically, current interactions with LLMs are based on turn-based conversation, where users input a complete query and wait for the model's full response before continuing. As shown in Figure~\ref{fig:duplex-1}, this conventional design relies on explicit end-of-turn signals to manage turn-taking behavior (such as typing ``enter'' in the chat interface), creating a rigid back-and-forth pattern where user input is disabled during model generation. This approach, while functional for text-based interactions, is unsuitable for fluid speech conversations that lack such explicit markers. The challenge lies in developing systems that can process partial inputs in real time, anticipate user intent, and seamlessly integrate into dynamic conversations without rigid structures. Such models must dynamically determine when to generate a response and when to continue listening, emulating natural conversation flow. Achieving this dynamic turn-taking behavior is crucial for creating natural, fluid conversations and avoiding awkward pauses or overlapping speech.

In a natural conversation, participants should be able to speak at any moment, just as humans do in face-to-face interactions. This creates two critical challenges in duplex dialogue systems: determining when to start talking (either when detecting the user has finished their turn or after a period of mutual silence) and deciding when to stop (either due to user interruption or completion of the system's response). As shown in Figure~\ref{fig:duplex-2}, we address these challenges through a dynamic end-of-turn detection method that processes speech input in chunks. For each incoming chunk, our system employs an automatic checking process that leverages the end-of-turn token inherently present in Llama3. This allows us to detect turn completion without requiring any specific training. Figure~\ref{fig:duplex-3} illustrates this checking process in detail - the system feeds accumulated chunks of input (from ``Tell me'' to ``Tell me a story about you'') into the model to obtain next-token probability distributions. When the probability of the end-of-turn token exceeds a predetermined threshold, the system interprets it as a completion signal and initiates response generation. Following human conversation etiquette, our system is designed to be polite and attentive - if the user interrupts during the system's generation, the model immediately stops its current generation to process the new input, ensuring the user always has priority in the conversation. Additionally, we implement a silence counter to track periods when neither party is speaking. When the silence exceeds a significant amount of time, the model initiates conversation to maintain engagement, similar to how humans naturally fill conversational gaps. 
In conclusion, we present a duplex interaction system that enables fluid conversation between users and systems. Our approach leverages Llama3's inherent capability and rule-based method to achieve natural turn-taking behavior without requiring additional specialized training.

\section{Data Preparation}  
\label{sec:data}

\subsection{Real Data Preparation} 
\label{subsec:realdata}
We used an in-house dataset with 40,000 hours of real-world speech data. To transform the raw real-world audio into structured conversations, we employed several speech-processing techniques to segment the conversation, and then we utilized an ASR model to obtain the corresponding transcriptions of the real-world audio. 

\subsubsection{Conversation Segmentation} 
Initially, speaker diarization is applied to determine the precise time intervals during which each speaker is active. Subsequently, source separation is utilized to resolve overlapping speech segments involving two or more speakers. Following this, the speech segments attributed to each speaker are concatenated by speaker verification to construct a continuous speech track for each individual. Finally, we segment the resulting speech tracks into conversations, ensuring a maximum of 25 conversational turns involving two speakers within a duration of five minutes.

For speaker diarization, we employ the off-the-shelf speaker diarization model~\citep{Plaquet23} available on Hugging Face\footnote{pyannote/speaker-diarization-3.1}. The source separation task is performed using a finetuned Sepformer model~\citep{sepformer}, which is trained on the Libri2mix dataset at a 16 kHz sampling rate. For speaker verification, we utilize the ECAPA-TDNN model~\citep{DesplanquesTD20}, which is available on Hugging Face\footnote{speechbrain/spkrec-ecapa-voxceleb~\citep{speechbrain}}, which has been trained using the VoxCeleb1~\citep{VoxCeleb1} and VoxCeleb2~\citep{VoxCeleb2} datasets.

\subsubsection{Transcription Generation}
To generate high-quality transcriptions of a large web-crawled Taiwanese-accented Mandarin speech corpus, we balance effectiveness and efficiency by utilizing the $\mathbf{K}^\mathbf{2}\mathbf{D}$ framework~\citep{tseng2024leave}, which employs knowledge distillation along with pseudo-labeling and filtering to enhance performance. Following this framework, we train a small Whisper~\citep{radford2023robust} model on the web-crawled Mandarin speech dataset due to its robust capabilities in handling diverse accents and noise, making it well-suited for our specific corpus. 

We further improve inference speed through the Faster-Whisper toolkit\footnote{https://github.com/SYSTRAN/faster-whisper}, a CTranslate2-based~\citep{klein-etal-2020-opennmt} reimplementation of OpenAI's Whisper model. To address hallucinations during silent periods of speech, we integrate Silero VAD~\citep{silerovad}, a pre-trained Voice Activity Detector. Additionally, we implement rule-based filtering to mitigate common hallucination patterns and improve overall transcription accuracy.

\subsection{Synthetic Data Preparation}
\label{subsec:synthetic}
In the preliminary study, we found that training a spoken LLM solely with real-world dialogue data causes severe catastrophic forgetting of the skills present in the initialized text LLM. When trained only on real-world dialogue, the spoken LLM cannot produce semantically coherent responses. This issue likely arises because real-world dialogue has a distribution that is very different from the training data distribution of initialized text LLM. To prevent catastrophic forgetting of the initial text LLM's skills and to expand the size of our training dataset, we used text LLMs to produce multi-turn text conversations and then transformed text dialogues into multi-turn spoken dialogues using our Text-to-Speech system.

\subsubsection{Creation of Text Dialogue}
We adopted strong text-based LLMs to generate diverse synthetic dialogue. Our synthetic dialogues can be classified into two types: ``with'' or ``without'' interruption and speaker overlapping. The former aims to simulate real-world full-duplex conversations in which people may interrupt each other or speak simultaneously, while the latter simply stands for turn-taking communication without any interruption or speaker overlapping. This helps us better simulate real-world conversation dynamics, where people either take turns speaking or talk over each other.

\begin{figure}[t]
    \centering
    \includegraphics[width=0.95\linewidth]{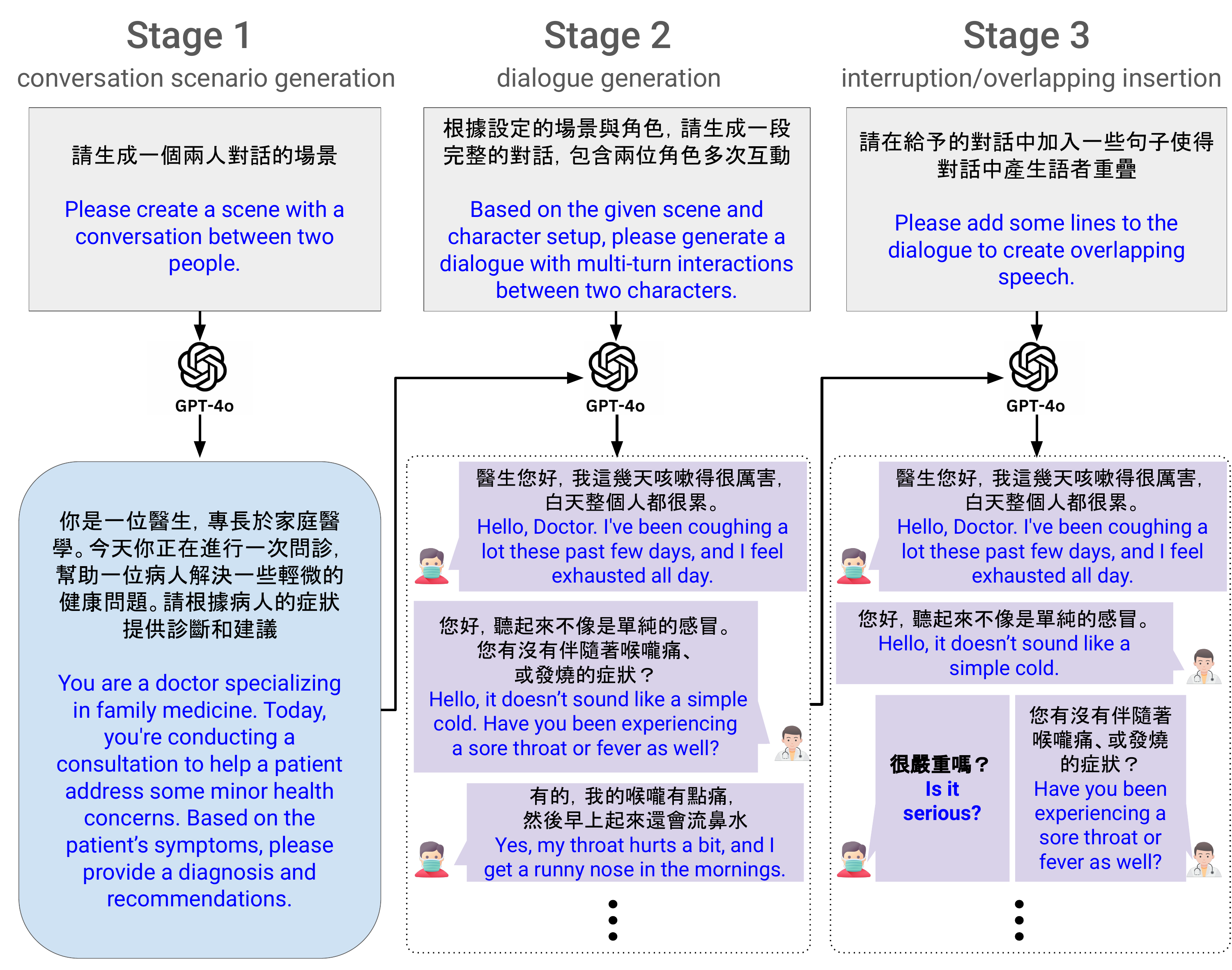}
    \caption{An illustration of the textual data generation pipeline. In Stage 3, the sentence in bold represents the inserted text, creating the simultaneous speech content. To save space, we simplified the prompt in the figure. The \textcolor{blue}{blue} text is the English translation of the corresponding Mandarin text.}
    \label{fig:textual-data-generation-pipeline}
\end{figure}

To generate high-quality and diverse synthetic dialogues, we developed a pipeline consisting of three stages: ``conversation scenario generation'', ``dialogue generation'' and ``interruption and overlapping insertion'',  as shown in Figure~\ref{fig:textual-data-generation-pipeline}. We first generated diverse dialogue scenarios as the basis of the subsequent steps in the first stage. Next, we produced turn-taking dialogues based on these scenarios, collecting and formatting them as training data without interruptions or overlaps. Finally, in the last stage, we inserted interruptions and overlapping speech into the dialogues to create training data that simulates full-duplex conversations. As a result, we obtained two distinct types of synthetic dialogues. Below, we provide the details of our pipeline.

In the conversation scenario generation stage, we prompted GPT-4o~\citep{hurst2024gpt} to generate a diverse set of scenarios, i.e., the role-playing settings, of the conversations. This led to approximately 200 distinct scenarios representing various everyday conversational contexts, such as interactions between users and AI assistants, doctors and patients, teachers and students, and wedding consultants and newlyweds, among others.


Next, in the dialogue generation stage, we produced turn-taking dialogues based on the scenarios generated in the previous stage. We employed two distinct generation strategies here. The first strategy is similar to Magpie~\citep{xu2024magpie}, a self-synthesis method that automatically generates both instructions and responses. Using this framework, we created multi-turn conversations in a turn-by-turn manner. Specifically, we generated a single turn with each instance of LLM inference, aggregating the already generated content to serve as the prefix and context for subsequent inferences. By utilizing a high temperature of 0.9, we found that the dialogues generated using this strategy were significantly diverse while maintaining coherence. For this approach, we selected LLaMA-3.1 70B due to its accessible tokenization process and strong generation capabilities. The second strategy involved handcrafting a system prompt to direct the model to function like a playwright, resulting in the generation of the entire multi-turn dialogue. For this approach, we selected GPT-4o, which demonstrated impressive abilities in producing consistent and coherent dialogues based solely on the system prompt. The dialogues generated with these strategies in this stage were collected and formatted to be the training data that did not contain interruption or speaker overlapping.

Finally, we turned the dialogues generated in the previous stage into full-duplex ones in the last stage. We again employed GPT-4o here. We provided the previously generated dialogues to GPT-4o and asked it to insert suitable interruptions or simultaneous speech contents into the existing dialogues. As part of the system prompts, a hand-crafted guideline was also provided to improve its performance. The guideline included several rules: 
\begin{enumerate}
    \item The inserted content should be coherent with the original dialogue, which means the model should be aware of the context and should not insert irrelevant content or violate the role-playing settings.

    \item The model should be sensitive to the speaker of the turn where it inserts content, and avoid speaker inconsistency, such as making someone interrupt himself/herself, etc.

    \item If inserting text into the existing conversation would result in awkward or unclear expressions, the model is allowed to make slight adjustments to the original dialogue to maintain fluency and coherence.
\end{enumerate}
With some demonstrations of each rule in the guideline, GPT-4o typically could insert natural and coherent interruptions or speaker overlapping into existing dialogues. In this stage, we collected the generated dialogue to serve as the full-duplex conversation data for training.

\subsubsection{Speech Synthesis} 
Once the text-based conversations are generated by the LLM, we employ TTS models to transform them into speech. 
However, there is currently no TTS model capable of producing natural-sounding speech in Taiwanese Mandarin. 
To address this, we fine-tuned CosyVoice-300M~\citep{du2024cosyvoice} using our in-house dataset, enabling it to generate speech in Taiwanese Mandarin. 
There will be another technical report available soon about the efforts to fine-tune CosyVoice.
For both instructions and responses in these conversations, we utilized our fine-tuned TTS model, randomly selecting pairs of utterances to represent different speakers in the dialogue.

\subsection{Interleaved Sequence Forming}
We transform spoken utterances into interleaved sequences of discrete text and speech units through a three-step process. First, speech units are extracted using a streaming HuBERT model, as detailed in Section~\ref{sec:streaming_unit}. Then, we transcribe the speech into text using Whisper with word level timestamp. Finally, texts are inserted into the speech unit sequence by the corresponding timestamps, as outlined in Section~\ref{sec:real-time-interleaving}.

\subsection{Data Filtering}
\label{sec:filtering}
We performed rule-based filtering to filter out low-quality data. The process focused on filtering out the data with the following patterns:
\begin{enumerate}
    \item \textbf{ASR Hallucination Filtering}: Since we used the Whisper model when forming the interleaved sequences of text and speech, hallucination may occur during this process. We identified some common hallucination patterns and filtered out the data containing these patterns.

    \item \textbf{Language Filtering}: As we aimed to develop a spoken LLM for Taiwanese Mandarin, we filtered out the data that were entirely in English in our real-world speech data.

    \item \textbf{Missing Turn Filtering}: We observed that some real-world multi-turn dialogue data may contain missing turns. This often happens during conversation segmentation, where parts of a dialogue might be unintentionally removed, causing breaks in the natural flow. Additionally, filtering out inaccurate or hallucinated words and sentences during transcription can create further gaps, making the data less suitable for training. To avoid these issues, we chose to exclude such data from our dataset.

\end{enumerate}

After the filtering, we processed the dataset for training, which will be explained in Section~\ref{sec:training}.

\section{Training}
\label{sec:training}
We describe the training setup of the components of our spoken LLM, including the LLM backbone and the speech decoder. 

\subsection{LLM Backbone Training} 
The training process for our spoken LLM consists of two primary phases: Pre-training and Supervised Fine-Tuning (SFT). During pre-training, the model learns to align text and speech units, which were unseen for the backbone model before, through tasks such as Automatic Speech Recognition (ASR) and Text-To-Speech (TTS). The SFT phase then focuses on refining the model's ability to handle spoken dialogue.

As previously mentioned in Section \ref{subsec:llm_backbone}, we aimed to equip the model with modality control ability. Thus, we processed the data in Section \ref{subsec:realdata} and \ref{subsec:synthetic} into various input/output modalities and generated corresponding system prompts, with the format like the one in Section \ref{subsec:llm_backbone}, to form the training dataset of the subsequent pre-training and SFT phases. We detailed the two training phases in the following sections.


\subsubsection{Pre-training}
\label{sec:pretraining}

Since we use a text-based LLM, i.e., LLaMA-3.1 8B, as the initialization of the spoken language model, the speech units are new to the text-based backbone. Therefore, it is crucial for the model to get familiar with these newly added speech units and the relationship between these units and the existing text tokens before the training of spoken dialogue. This is the goal of the pre-training phase.

The pre-training phase utilizes a diverse dataset comprising 24k hours of speech data. In this stage, the model is trained to align text tokens with speech units through ASR and TTS tasks. For ASR, the model learns to recognize the text content of the speech based on a sequence of speech units and predict the text content or the interleaved sequence of text content and the speech units (hybrid modality). We did not include the hybrid modality as the input modality because this will make the task too simple to build the text-speech alignment, i.e., the model only learns to extract the text tokens from the interleaved sequence without any understanding of alignment. Nevertheless, we incorporated the hybrid modality for the labels in the hope of strengthening the text-speech alignment, since the model had to know which speech units in the given sequences represent each predicted text token. 

On the other hand, TTS involves training the model to generate speech units corresponding to the given text content. Here, we defined the so-called ``correspondence'' as follows: a unit sequence corresponds to the text content if and only if the voice synthesized with that unit sequence can clearly convey the content. While the ASR task is discriminative, the TTS task is essentially generative. These two kinds of training can complement each other to help the model learn the alignment better. Similar to the ASR training, we include the hybrid, as well as the pure units, as a possible label modality to enhance the text-speech alignment.

The training data for the aforementioned two tasks originated from the real speech data in Section~\ref{subsec:realdata} and the synthetic dialogue data in Section~\ref{subsec:synthetic}.
Specifically, we drew a subset of the prepared speech data and separated all the turns in those multi-turn dialogue data into individual single-turn data. We then used the prepared transcriptions and the units to form the datasets for ASR and TTS. We also included the system prompts containing corresponding input/output modality combinations and the role-playing instructions with the ``speech recognition model'' and ``text-to-speech model'' as the roles.

Besides ASR and TTS, this stage also included text-based training data to ensure that the model retains its language comprehension abilities. 
The text-based training data consists of the transcriptions of the multi-turn spoken dialogue data (without interruption or speaker overlapping) that are approximately equivalent to 20k hours of speech, resulting in a pre-training dataset totaling about 44k hours of speech. 
The equivalent duration of the purely textual data was determined by their speech counterpart. The input consists of a system prompt that incorporates both the modality control prefix, where both the input and output modalities are text tokens, and the role-playing instructions that define the roles of the model depending on the scenarios of the dialogues. We used the original multi-turn conversation template of LLaMA-3.1 to concatenate the turns, with special tokens of headers inserted between turns and turns to specify the current speaker for each turn. The headers will follow the pattern: $<$$|\mathtt{start\_header\_id}|$$>$ $\mathtt{speaker}$ $<$$|\mathtt{end\_header\_id}|$$>$, where the $\mathtt{speaker}$ could be replaced with the exact speaker role of the current turn. The model is trained to predict the entire multi-turn conversation. 

The total amount of the data and the input/output modalities of all the tasks in the pre-training phase are summarized in Table~\ref{tab:pretrain_data}.


\begin{table}[ht]
    \centering
    \begin{tabular}{c c c c}
        \toprule
        \textbf{Task} & \textbf{Input} & \textbf{Output} & \textbf{Estimated duration (hr)}\\
        \midrule
        \multirow{2}{*}{ASR} & \multirow{2}{*}{unit} & text & 6k \\
        {} & {} & hybrid & 6k \\
        \midrule
        \multirow{2}{*}{TTS} & \multirow{2}{*}{text} & unit & 6k \\
        {} & {} & hybrid & 6k \\
        \midrule
        Text-based & text & text & 20k \\
        \bottomrule
    \end{tabular}
    \caption{The input/output modalities and the estimated durations of the tasks for pre-training.}
    \label{tab:pretrain_data}
\end{table}

\subsubsection{SFT}
\label{sec:sft}

The SFT phase focuses on enhancing the model's proficiency in multi-turn conversation with various input/output modalities and role-playing scenarios. This stage involves fine-tuning the model on an extensive dataset with a total duration of $\sim$100k hours, ensuring the model can effectively understand and communicate in speech in diverse contexts and scenarios. 

To preserve the model's text-speech alignment capabilities developed in the pre-training phase, ASR tasks with 12k hours of speech data used in the pre-training were also incorporated into the SFT phase. This ensures the model retains its proficiency in mapping between text and speech while improving its conversational abilities. 

We included multi-turn dialogue data with various input/output modality combinations in the SFT phase. Purely text-based dialogues, i.e., input and label are both in text, were employed to maintain the model's general textual ability. Here, we only used the dialogue data from Section~\ref{subsec:realdata} and \ref{subsec:synthetic} that did not contain any interruption or speaker overlapping for purely text-based training data. The text-based data equivalently amounted to about 29.5k hours with appropriate processing like the one for the text-based pre-training data.

As for the spoken dialogue data that involved speech units, we used 55.5k hours of spoken dialogues. We restricted the possible input/output modality combinations for these data into only five different categories, with three kinds of input modalities (pure text tokens, pure speech units, and interleaved sequence of the text tokens and speech units) and two kinds of output modalities (pure text tokens and hybrid). All the possible combinations were included except for the combination with input and output modalities both in text. The output modality of pure speech units was excluded because we hope that the model can respond with speech while simultaneously outputting the corresponding text content for the user to view, making the pure speech unit sequence unsuitable here. On the contrary, The purpose of including the pure units as one of the input modalities was to make the model familiar with conversations whose inputs contain no text tokens, thus mitigating the reliance on the streaming ASR model in Figure~\ref{fig:framework}. Ideally, this can have benefits in that we can omit the ASR and directly input the extracted units into the spoken LLM, reducing the latency and allowing more real-time interaction.

We processed the transcriptions and the extracted unit sequences of the data to formulate the dialogues into an equal amount for each input/output modality combination. The process was similar to the one for the pre-training dataset but with two distinctions. The first one was that we did not restrict the conversation data to be single-turn like what we did for ASR and TTS training. Instead, we formulated them with the inherent multi-turn template of LLaMA-3.1 and used the header tokens to specify the speakers of the turns, which was similar to the processing of the text-based dialogues. 

\begin{figure}
  \centering
  \includegraphics[width=0.8\linewidth]{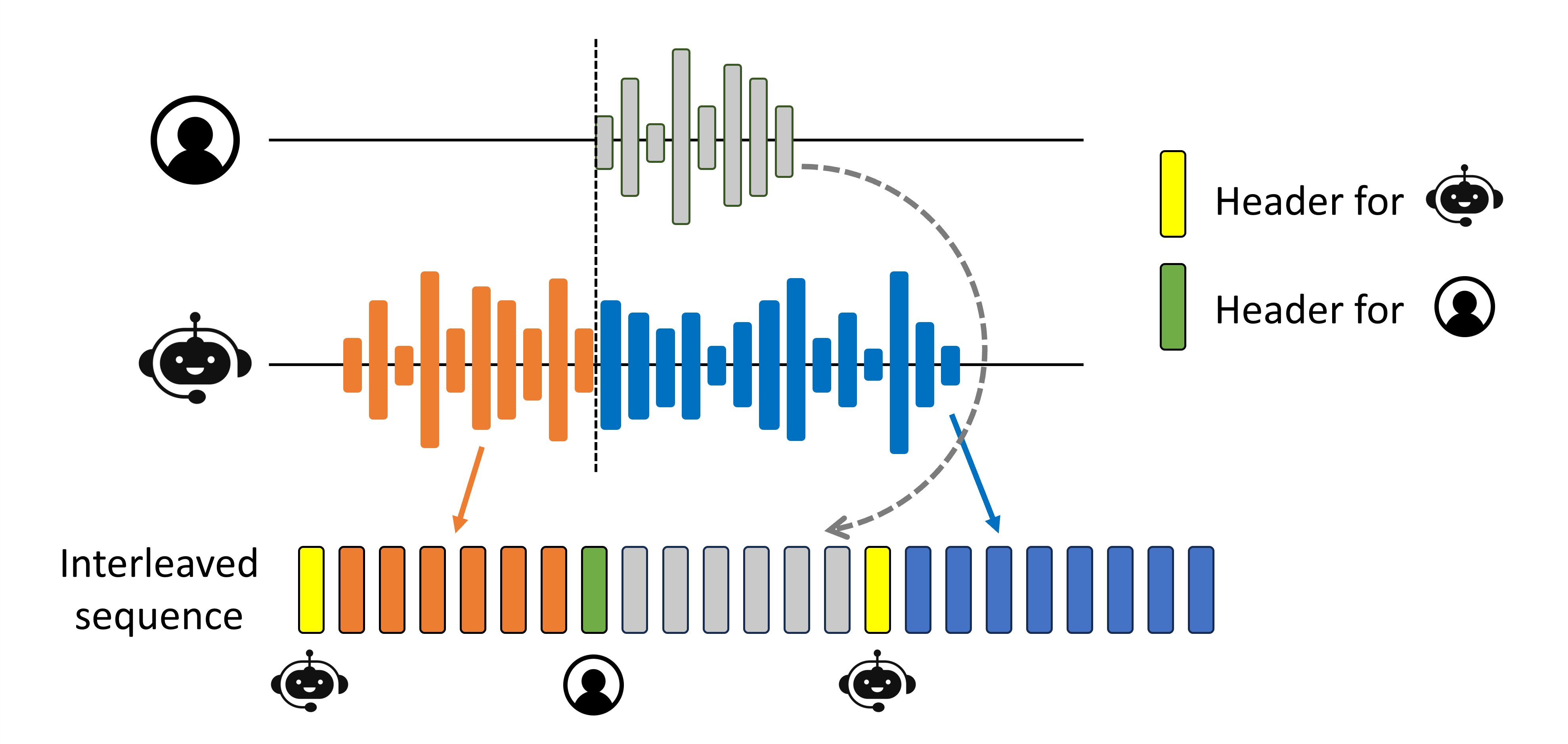}
  \vspace{-0.5em}
  \caption{An illustration of how we represent the speaker overlapping or interruption in the input sequences of the model.
  \label{fig:interrupt}}
  \vspace{1em}
\end{figure}

The other distinction was that we included the case of interruption and speaker overlapping in the training data of spoken dialogues, which was excluded for purely text-based dialogue training. We again used the header tokens to mark the speaker changes, and we inserted the interrupting content at the point where the current speaker's speech was interrupted. This divided the turn of the current speaker into two parts and made it incomplete as a sudden speaker switching occurred. An illustration of this processing is demonstrated in Figure~\ref{fig:interrupt}. Although this method compressed the two simultaneous speech signals into a single stream, the model should still learn to predict the second part of the content despite the interruption. We believe this can enable the model to know what it is saying and what it intends to say next, even when someone interrupts it or speaks at the same time.

We generated a diverse set of system prompts depending on the input/output modality combinations and the context of the dialogues, following a similar methodology for the pre-training data. The total amount of the data and the input/output
modalities of all the tasks in the SFT phase are summarized in Table~\ref{tab:sft_data}.


\begin{table}[ht]
    \centering
    \begin{tabular}{c c c c} 
        \toprule
        \textbf{Task} & \textbf{Input} & \textbf{Output} & \textbf{Estimated duration (hr)}\\
        \midrule
        \multirow{2}{*}{ASR} & \multirow{2}{*}{unit} & text & 6k \\
        {} & {} & hybrid & 6k \\
        \midrule
        Text-based & text & text & 29.5k \\
        \midrule
        \multirow{6}{*}{Spoken Dialogue} & \multirow{2}{*}{unit} & text & 11.1k \\
        {} & {} & hybrid & 11.1k \\
        \cmidrule{2-4}
        {} & \multirow{2}{*}{hybrid} & text & 11.1k \\
        {} & {} & hybrid & 11.1k\\
        \cmidrule{2-4}
        {} & text & hybrid & 11.1k\\
        \bottomrule
    \end{tabular}
    \caption{The input/output modalities and the estimated durations of the tasks in SFT phase.} 
    \label{tab:sft_data}
\end{table}

\subsection{Setup}
The pre-training and SFT were conducted with the NeMo toolkit~\citep{kuchaiev2019nemo}. Due to the memory issue, we truncated the dialogues into a maximum length of 16k tokens and units per dialogue sample. In the pre-training phase, we trained the model on 16 H100 GPUs with a learning rate of 1e-5 and a batch size of 64. The pre-training takes 65,000 steps with 2\% of the steps for warmup, taking 960 H100 GPU-hours in total. Regarding the SFT phase, the training was done with a learning rate of 1e-5 and a batch size of 128. The process took 32 H100 GPUs and 20,000 steps, resulting in about 2300 H100 GPU-hours for the entire training. The same warmup strategy as the one used during pre-training was applied.

For decoder training, we use a single-speaker dataset on Hugging Face\footnote{https://huggingface.co/datasets/voidful/gen\_ai\_2024}. In this stage, we train the model on 8 
NVIDIA-V100 GPUs, with a learning rate of 2e-4 and a batch size of 512 per device. The entire training process took around 600 V100 GPU-hours.

\section{System} \label{sec:system} 
In this section, we elaborate on how to utilize different components to build a streaming pipeline. Section~\ref{sec:components} introduces each streaming component, Section ~\ref{sec:optim} introduces optimizations applied in our streaming pipeline, and finally in Section~\ref{sec:latency}, we discuss the latency of the system. Our code is publicly available\footnote{https://github.com/nervjack2/SpeechChatGPTStreaming/tree/end2end}.

\subsection{System Components}
\label{sec:components}
\subsubsection{Streaming ASR}
\label{sec:streaming_ASR}
For streaming ASR, we follow~\citet{machavcek2023turning}, converting Whisper into a streaming version. 
Specifically, during real-time inference, we send an audio segment to Whisper every 0.3 seconds to get hypothesis transcriptions. 
Then, we compare the current segment's hypothesis transcription with those of previous segments to see if there are identical prefixes. 
If there is an identical prefix, that prefix becomes a confirmed transcription.
All the other parts remain in the audio buffer to continue matching prefixes.

In the original paper, there are two different methods for how to trim the confirmed transcriptions. 
The first method trims complete sentences, while the second one trims a fixed number of seconds. 
The drawback of the first method is that if the sentence is too long, it can lead to long latency during real-time inference. 
Therefore, we choose to use the second method.

Based on previous work, we have also implemented some new features: 
\begin{enumerate}
\item \textbf{Warmup}: Our system feeds a random audio clip to Whisper for a warmup before any client connections are made.
\item \textbf{De-Hallucinate}: We use a rule-based approach to eliminate hallucinations by filtering out some commonly hallucinated patterns.
\item \textbf{Reset}: We have implemented a feature to reset the entire ASR system in real time. It resets when there is a hallucination, turn taking, or when the user types the special command `===' with text.
\end{enumerate}

\subsubsection{Streaming Speech Units}
\label{sec:streaming_unit}
\begin{figure}
  \centering
  \includegraphics[width=0.7\linewidth]{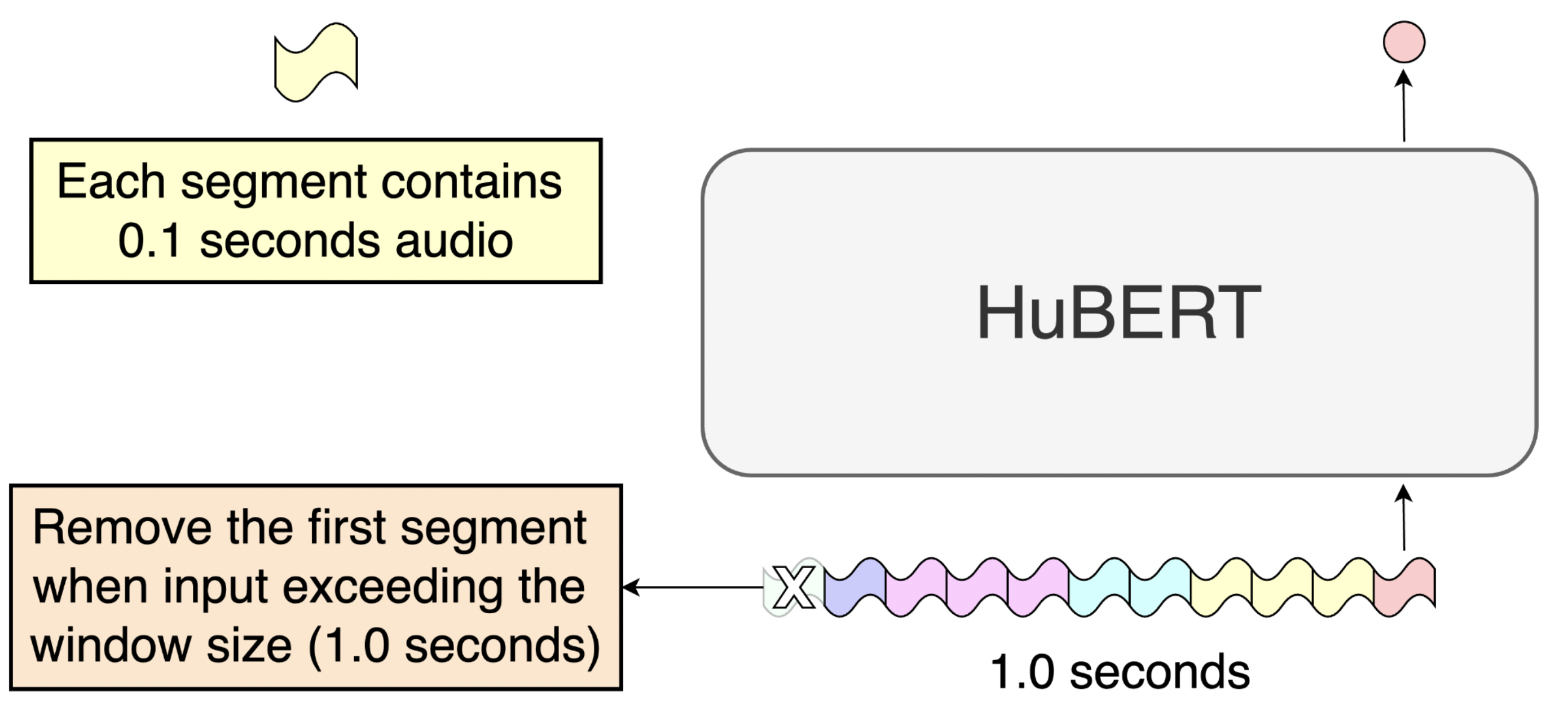}
  \vspace{-0.5em}
  \caption{An illustration of how we extract HuBERT units during real-time inference streamingly.
  \label{fig:streaming-speech-units}}
  \vspace{1em}
\end{figure}

When extracting speech units in real-time, as shown in Figure~\ref{fig:streaming-speech-units}, we divide the input audio into segments, each containing 0.1 seconds of audio. 
Whenever we receive a new 0.1-second audio clip, we add it to the audio buffer and then send all the audio segments from the buffer to HuBERT~\citep{hsu2021hubert} to extract speech units. 
We extract speech units for the new 0.1-second audio clip, conditioned on all the audio segments in the audio buffer.
To prevent the audio length from increasing indefinitely, a window size of 1 second is applied. 
When the total length of the audio in the buffer exceeds this window size, we discard the oldest audio segment from the buffer.

\subsubsection{Real-time Interleaving}
\label{sec:real-time-interleaving}
\begin{figure}
  \centering
  \includegraphics[width=0.85\linewidth]{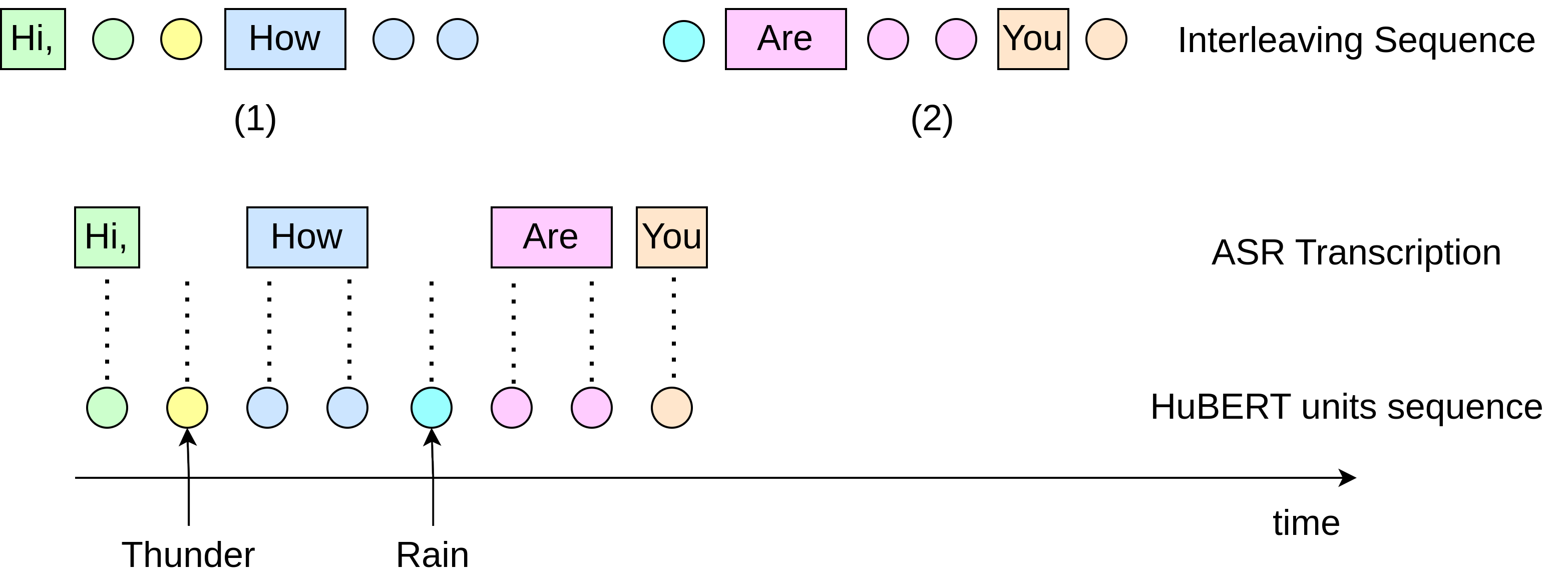}
  \vspace{-0.5em}
  \caption{An illustration of interleaved transcription and speech units in real-time. We use a simple toy example as an illustration, where the ASR model first generates (1) Hi, How, and then generates (2) Are You. Here, ``Thunder'' and ``Rain'' refer to the background environmental sounds in the input speech.
  \label{fig:interleaving}}
\end{figure}

In real-time, we need to interleave transcriptions and their corresponding speech units before sending them to language models. 
Since the latency of ASR is much longer than that of HuBERT, we choose to perform interleaving around the transcriptions.
We present a simple toy example for illustration, where the ASR model generates (1) Hi, How, followed by (2) Are You.
Specifically, as shown in Figure~\ref{fig:interleaving}, whenever ASR produces a new transcription, we obtain the time spans for different words within that transcription. 
Based on these time spans, we take the corresponding sequences of speech units for interleaving. 
This method of interleaving might discard some speech units between two transcriptions, for example, the light blue units in Figure~\ref{fig:interleaving}.
To address this issue, we check for any speech units that were missed between the current and the previous transcriptions and append up to 2 seconds of these missed speech units to the beginning of the interleaved sequence.

\subsubsection{Streaming Speech Unit Decoder}
In Section~\ref{subsec:decoder}, we describe a diffusion-based decoder trained to synthesize output speech units back into audio. However, this decoder is designed for offline processing and does not inherently support streaming. To approximate streaming in the inference phase, we apply chunking, similar to the streaming approach in ASR, where speech unit segments are sent to the decoder in real time. Despite this, the audio generated by the decoder suffers from discontinuities at segment boundaries, affecting overall quality. Replacing it with a fully functional streaming decoder is left for future work.

Although directly fitting an offline decoder into a streaming scenario is far from perfect, we still try to mitigate the artifact by decreasing the number of total chunks. We propose a \textit{dynamic chunking} strategy which aims to make each chunk as long as possible without increasing the latency.

\begin{figure}
  \centering
  \includegraphics[width=1.0\linewidth]{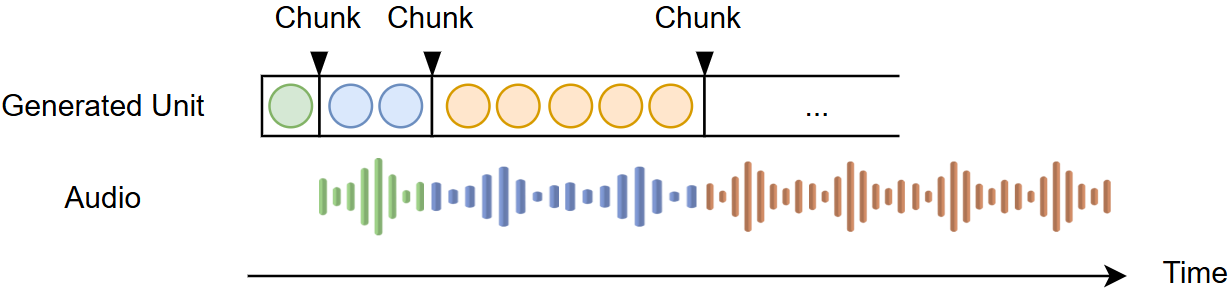}
  \caption{The dynamic chunking strategy.
  \label{fig:dchunk}}
\end{figure}

Figure~\ref{fig:dchunk} illustrates the dynamic chunking strategy. We always chunk as late as possible, before the audio resynthesized from the previous chunk is finished playing. If the speech units are generated faster than real-time, we can expect the $k$-th chunk to be longer than the $(k-1)$-th chunk, overall decreasing the number of chunks needed and therefore the occurrence of the boundaries. Since we can calculate the time represented by each chunk in advance, we can schedule the next chunk as late as possible. Denote the time of the $k$-th chunking as $t_k$, the size of the $k$-th chunk as $n_k$, and the time represented by each unit as $t_{unit}$. The next chunking is scheduled as
\[
t_{k+1} = t_k + n_k\times t_{unit} - \epsilon ,
\]
where one can adjust $\epsilon$ manually to resynthesize the next chunk slightly earlier. Since decoder forwarding requires processing time, the design aims to prevent gaps between chunks by ensuring the next chunk is resynthesized before the previous audio chunk finishes playing.

\subsection{System Optimization}
\label{sec:optim}
\subsubsection{Turn-Taking Implementation}
\label{subsec:tti}
\begin{figure}
  \centering
  \includegraphics[width=1.0\linewidth]{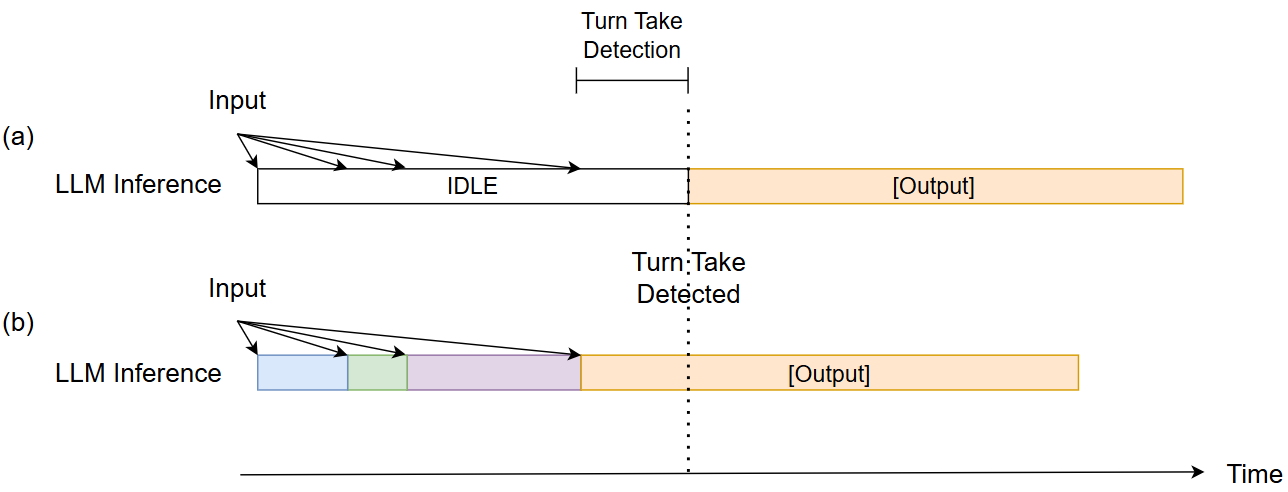}
  \caption{Two different turn-taking implementations. (a) The straightforward implementation. (b) The optimized asynchronous version without idle.
  \label{fig:optim}}
\end{figure}

Unlike the offline scenario, in the streaming case, LLM requires a turn-taking protocol to determine when the user stops and starts the inference session. The determined process itself also introduces latency, as shown in Figure~\ref{fig:optim}a.

To mitigate the latency, note that the determined process and inference session can be performed asynchronously. The key idea is to restart a new inference session immediately after receiving any new input. The output is only confirmed when a turn-taking signal is detected, as illustrated in Figure~\ref{fig:optim}b. In this implementation, the turn-taking detection process overlaps with the inference session, allowing the algorithm to operate without idle waiting.

\subsubsection{LLM Inference Acceleration}
Thanks to the rapid development of LLM in recent years, numerous frameworks exist to accelerate the inference speed of LLM. In this work, we use TensorRT-LLM\footnote{https://github.com/NVIDIA/TensorRT-LLM} to accelerate the inference speed. After applying TensorRT-LLM with int4 weight-only-quantization, our LLM can generate about 100 tokens per second on a single Nvidia GeForce RTX 3090 GPU. Since we use 25Hz speech units, the speed is fast enough for a real-time system. Note that without acceleration, our LLM can only generate about 20 tokens per second, which does not apply to building a real-time system.

\subsection{System Latency}
\label{sec:latency}
The latency is defined as the time between the user input ends and the system starts to speak. Streaming ASR with interleaving produces nearly 900ms latency, LLM needs about 400ms to generate the first token, and the decoder needs about 200ms. Additionally, the turn-taking wait time is capped at 1 second, resulting in a maximum latency of 2500 ms. By applying the technique described in Section~\ref{subsec:tti}, we reduce the maximum possible latency to 1900 ms.

Such latency is still very high compared to humans, which is reported to be under 500ms~\citep{jacoby2024humanlatencyconversationalturns}. One reason is that streaming Whisper and streaming decoder both apply a chunking algorithm, which inevitably introduces a large latency. Applying these models trained under offline scenarios for streaming harms their performance a lot and is also problematic. This suggests that all components should inherently support streaming when building a real-time system.

\section{Evaluation} \label{sec:evaluation} 

\subsection{Multi-turn Dialogue}

We design multi-turn dialogues on various topics, including expressing opinions on hip-hop music, daily discussions about pet care, casual movie chats, thoughts on social media, and relationship counseling. 
These topics range from everyday conversations to more specialized discussions. The objectives of the evaluation process are:

\begin{enumerate}
    \item \textbf{To test the model's ability to understand the user's speech and respond appropriately within the topic’s context.} 
    This is because the ability to respond appropriately within the context of the topic is a fundamental metric for evaluating the model’s conversational skills. 
    Our goal is to maintain the original text language model’s proficiency in handling text-based content.

    \item \textbf{To evaluate the intelligibility of the model’s spoken responses.}
    The model's speech output forms the user's first impression of its performance. 
    Even if the textual content is appropriate, low intelligibility or unnatural speech could significantly impact the user’s perception and cognition of the model's quality.

\end{enumerate}

Our methodology for constructing multi-turn dialogues involves two steps: first, selecting representative topics as conversation scenarios, followed by having human experts design structured user utterances to facilitate model interactions. 
For example, we might select ``restaurant reservation'' as a scenario, where human experts design sequential user inputs such as asking about table availability, specifying dining preferences, and confirming reservation details. 
Similarly, for a ``technical support'' scenario, the designed dialogue might include users describing their problems, responding to troubleshooting questions, and acknowledging solution steps.
Table~\ref{tab:multi-turn-dialogues} presents a sample dialogue of expressing opinions on hip-hop music, with additional examples provided in Appendix~\ref{appendix:multi-turn-dialogue-example}.
To evaluate the model's capabilities in handling both text and speech within conversations, we propose several methods as detailed in the following sections.

\begin{longtable}{p{40em}}
    \caption{The system prompt tells the model what role it should play. The modality control is omitted for simplicity. \textbf{User} refers to the input from the user, and \textbf{Machine} refers to the model's response. 
    The user's input is speech, represented here by its text transcription. 
    The model's output is in an interleaved format of text and discrete speech units, with only the text part shown here. 
    The \textcolor{blue}{blue} text is the English translation.
    } \\
    \toprule
    \\
    \texttt{System Prompt:\begin{CJK}{UTF8}{bkai}你是一個對於嘻哈音樂非常有研究的聊天機器人，精通各種背景知識，今天你正在和使用者進行一段有趣的對談。\end{CJK}} \\
    \texttt{\textcolor{blue}{You are a chatbot with deep knowledge of hip-hop music, mastering various background information. Today, you are having an interesting conversation with the user.}} \\ \\
    \midrule \\
    \texttt{User:\begin{CJK}{UTF8}{bkai}最近我發現了一個很棒的饒舌歌手，叫做熊仔。你可以和我介紹嘻哈音樂嗎？\end{CJK}} \\
    \texttt{\textcolor{blue}{Recently, I discovered a great rapper named Kumachan. Can you introduce me to hip-hop music?}} \\ \\
    \texttt{Machine:\begin{CJK}{UTF8}{bkai}大家好我是聊天機械熊仔是一位非常優秀的饒舌歌手他的音樂結合了臺灣的韻輸和西方的音樂元素令人讓人覺得非常新鮮雖然熊仔的音樂風格比較獨特但也創造出了許多讓人感動的歌曲像他最有名的作品就是他自己曾經演唱過的橋板上的女孩這首歌以淡微的旁白搭配著吉他符號創造出如此生動的影象同時呢也是讓大家對於這首歌充滿期待\end{CJK}} \\
    \texttt{\textcolor{blue}{Hello everyone, I'm your chatbot. Kumachan is an excellent rapper. His music combines Taiwanese rhyme with Western music elements, giving it a refreshing feel. Although his style is unique, it has produced many touching songs. One of his most famous works is *The Girl on the Bridge*, which blends subtle narration with guitar chords, creating vivid imagery that captivates listeners.}} \\ \\
    \texttt{User:\begin{CJK}{UTF8}{bkai}嘻哈音樂想表達的想法是標新立異、叛逆，還是對生活的一種反抗呢？你的看法是什麼？\end{CJK}} \\
    \texttt{\textcolor{blue}{Does hip-hop music express originality, rebellion, or resistance to life? What’s your opinion?}} \\ \\
    \texttt{Machine:\begin{CJK}{UTF8}{bkai}這就要看希哈音樂家個人的個性和創作意圖了因為每個人都有不同的感受和想法但是從一般來說習恨音樂可能更注重表達對未來的想像和夢想還有對社會問題的關切例如許多饒舌歌手通過音樂探討社會問題提倡改善或者用音樂表達對年輕人追求夢想的支持甚至有的歌手將音樂當成一種對政治和環境的批評和鬥爭所以其實習哈音樂的主題和方向非常多樣不能只用一致的樣子來描述\end{CJK}} \\
    \texttt{\textcolor{blue}{It depends on the artist’s personality and creative intent, as everyone has different feelings and thoughts. Generally, hip-hop music emphasizes imagining the future, dreams, and concern for social issues. Many rappers address societal problems, advocate change, or encourage young people to pursue their dreams. Some even use music as a means to critique politics and environmental issues. So, the themes and directions in hip-hop are highly diverse and can’t be confined to one description.}} \\ \\
    \texttt{User:\begin{CJK}{UTF8}{bkai}熊仔在自信這首歌中有提到最可怕的不是失去自信，最可怕的是你開始適應。你覺得可以怎麼解讀？\end{CJK}} \\
    \texttt{\textcolor{blue}{In Kumachan’s song "Confidence," he says, "The scariest thing is not losing confidence, but getting used to it." How would you interpret that?}} \\ \\
    \texttt{Machine:\begin{CJK}{UTF8}{bkai}我覺得最可怕的是沒有信心去追求自己的夢想即使失去一點小時,也不要放棄因為沒有事情能夠一次完成總是需要努力和堅持相信自己和夢想然後一步一步前進同時也要學會接納自己的情緒和經驗不要太快的否定自己這樣子的方式不僅能夠讓自己走得更近夢幻也能夠讓自己的人生變得更有趣和有意義\end{CJK}} \\
    \texttt{\textcolor{blue}{I think the scariest thing is not having the confidence to pursue your dreams. Even if you experience setbacks, don’t give up—nothing can be achieved in one attempt. It always takes effort and persistence. Believe in yourself and your dreams, and move forward step by step. At the same time, learn to embrace your emotions and experiences without being quick to deny yourself. This approach will not only bring you closer to your dreams but also make life more meaningful and enjoyable.}} \\ \\
    \texttt{User:\begin{CJK}{UTF8}{bkai}最後用嘻哈的風格來饒一段：這不是那種給你好好聽的，這是那種給我好好聽著，我不是那種一直在那講的，我都直接走上台拿獎的。\end{CJK}} \\
    \texttt{\textcolor{blue}{Finally, freestyle a rap in hip-hop style: "This isn’t something you listen to nicely—it’s something I tell you to listen to carefully. I’m not the one who just talks—I go straight on stage and win awards."}} \\ \\
    \texttt{Machine:\begin{CJK}{UTF8}{bkai}如果說又飛的話就會有一隻鞋子穿進去然後你就會被文字包住之類的這就是我們現在正在做的事情只不過除了你跟我聊天這件事情我現在已經不是這麼簡單了因為現在我要做的事情並不是隻有回答問題而是要和你聊天和分享一些更有趣的東西像是今天我們正在聊的嘻哈音樂這種類型的題目除了我剛剛跟你描述的原因之外我還想跟你分享一些嘻哈音樂的故事和相關知識等等甚至是有沒有什麼你想要知道的有關嘻哈音樂的問題我也都歡迎你問我\end{CJK}} \\
    \texttt{\textcolor{blue}{If I were to fly, a shoe would slip in, and you’d be wrapped in words or something like that. That’s what we’re doing right now. But it’s more than just chatting with you—it’s about sharing even more interesting things, like today’s discussion on hip-hop music. In addition to the reasons I’ve already given, I’d love to share more stories and knowledge about hip-hop. If you have any questions about hip-hop music, feel free to ask me.}} \\ \\
    \bottomrule
    \label{tab:multi-turn-dialogues}
\end{longtable}

\subsubsection{Contextual Coherence and Accuracy Evaluation via LLM Score}
In order to assess whether the model's responses align with the dialogue's context and situation, we employ the LLM-as-a-judge approach~\citep{zheng2023judging, chiang2023can}. 
Specifically, we provide GPT-4o\footnote{The version we used in this work is \texttt{gpt-4o-2024-05-13}.}~\citep{achiam2023gpt} with the dialogue topic and the entire conversation history. 
The goal of this method is to use large language models to predict whether the model's responses match the dialogue topic and maintain consistency with the conversation's context. 

The detailed evaluation prompt is presented in Appendix~\ref{appendix:prompt_template}.
The evaluation prompt presented in this work is designed to systematically assess the conversational performance of a voice agent. 
LLM Evaluators are instructed to rate the interaction between the agent and a human user across three specific dimensions: relevance and accuracy, adherence to the system prompt, and grammatical correctness. 
Each dimension is scored on a scale of 1 to 5, where 1 represents poor performance and 5 represents excellence.
    
The \textbf{Relevance and Accuracy} dimension aims to evaluate the agent's ability to respond appropriately to the user's requests. 
Evaluators are asked to consider whether the responses directly address the user's questions and provide correct, relevant information.
In the \textbf{Adherence to System Prompt} dimension, the focus shifts to whether the agent follows the instructions or role outlined in the system prompt. The evaluation ensures that the agent's responses remain consistent with its designated role and knowledge constraints.
The \textbf{Grammatical Correctness} criterion assesses the linguistic quality of the agent's replies, evaluating them for proper sentence structure, word usage, and overall fluency.
    
For each category, the LLM evaluator provides a brief justification explaining their rating. 
After rating all three dimensions, an overall score is calculated by averaging the three ratings and rounding to the nearest whole number.
The overall score represents the contextual coherence of the conversation on a scale from 1 to 5. The higher overall score indicates stronger contextual coherence.
This comprehensive scoring system enables a clear, structured evaluation of the voice agent's conversational capabilities across multiple critical aspects.

\subsubsection{Intelligibility Assessment}
In order to measure the model's intelligibility, we focus on its output, which alternates between text and speech units. 
We use the ASR model, Whisper\footnote{\url{https://huggingface.co/openai/whisper-large-v3}}~\citep{radford2023robust}, to transcribe the speech segments and then calculate the character error rate (CER) of these transcriptions, with the text parts extracted from the interleaved outputs of the model serving as the references. 
Since our model outputs in Taiwanese Mandarin, CER is used as an appropriate metric for accuracy, providing an objective measure of the model's ability to communicate effectively.

\subsubsection{Speech Quality Assessment via MOS Prediction Network}
We use a deep neural network-based model, NISQA\footnote{\url{https://github.com/gabrielmittag/NISQA}}~\citep{mittag2021deep}, to predict the mean opinion score (MOS) for the quality of the model’s speech output. 
The MOS is rated on a scale from 1 to 5, where 1 corresponds to the lowest quality and 5 to the highest. This score provides insight into the quality of the speech generated by the model.

\begin{table}[ht]
\setlength\tabcolsep{6pt} 
\renewcommand{\arraystretch}{1.0} 
\caption{Evaluation results. 
The \textbf{Model} field indicates different checkpoints throughout the training process. 
The \textbf{Modality} field indicates whether the input to the model includes text. 
\textbf{CER} stands for Character Error Rate, which measures the accuracy of the model’s output. 
\textbf{MOS} refers to the Mean Opinion Score, which is predicted using a DNN-based model. 
We report predicted average MOS score and 95\% confident interval.
\textbf{LLM Score} evaluates the model’s responses within conversations using large language models.} 
\centering
\begin{tabular}{c c c c c c}
\toprule
\textbf{Model} & \textbf{Modality} & \textbf{CER (\%) $\downarrow$} & \textbf{MOS $\uparrow$} & \textbf{LLM Score $\uparrow$} \\
\midrule
\texttt{spml-omni-it-mid} & s2s & 30.45 & 3.26$\pm$0.30 & 3.0 \\
\midrule
\texttt{spml-omni-it-mid} & u2s & \textbf{28.30} & 3.20$\pm$0.30 & 1.8 \\
\midrule
\texttt{spml-omni-it-last} & s2s & 28.99 & 3.42$\pm$0.32 & \textbf{3.4} \\
\midrule
\texttt{spml-omni-it-last} & u2s & 32.80 & \textbf{3.51$\pm$0.33} & 1.8 \\
\bottomrule
\end{tabular}
\label{tab:dialogue-results}
\end{table}

\subsubsection{Results}
\label{subsec:results}
We evaluate our spoken LLM at two critical stages of the training process: an intermediate checkpoint (\texttt{spml-omni-it-mid}) representing the model's performance midway through SFT training, and the final checkpoint (\texttt{spml-omni-it-last}) showing the model's ultimate capabilities. 
The results are presented in Table~\ref{tab:dialogue-results}.
The checkpoints \texttt{spml-omni-it-mid} and \texttt{spml-omni-it-last} correspond to SFT training steps 12864 and 19698, respectively.
We evaluate their spoken conversation proficiency in two kinds of modality settings: ``s2s'' for the case that the input from the user and the model's response are both in \textbf{hybrid} modality, and ``u2s'' where the user input is purely speech \textbf{unit} sequence while the model still responds in \textbf{hybrid} modality. The two modality combinations are employed to see whether our model has satisfactory ability when the streaming ASR in Figure~\ref{fig:framework} is not applied for the model's input.


The analysis of the results from the table reveals the following key observations:
\begin{enumerate}
    \item \textbf{CER (Character Error Rate)}: 
    Among the models, the lowest CER is observed for the \texttt{spml-omni-it-mid} in the \texttt{u2s} modality with a value of 28.30\%, indicating better intelligibility in this specific setting compared to others. 
    On the other hand, the \texttt{spml-omni-it-last} in the \texttt{s2s} modality shows a slightly higher CER of 28.99\%, which is still relatively low, suggesting decent performance in terms of intelligibility.

    \item \textbf{MOS (Mean Opinion Score)}: In terms of perceptual speech quality, the \texttt{spml-omni-it-last} in the \texttt{u2s} modality achieved the highest MOS of 3.51$\pm$0.33, indicating the best perceived quality of speech. 
    This is followed closely by the \texttt{spml-omni-it-last} in the \texttt{s2s} modality with a MOS of 3.42$\pm$0.32.

    \item \textbf{LLM Score}: For dialogue consistency, the highest LLM Score is observed in the \texttt{spml-omni-it-last} in the \texttt{s2s} modality, scoring 3.4, which suggests that the model maintained the best coherence with the conversation context. 
    In contrast, the lowest LLM Score of 1.8 is seen in both \texttt{u2s} modalities, implying lower consistency in these scenarios.

\end{enumerate}


In summary, \texttt{spml-omni-it-last} demonstrates compelling results in \texttt{s2s} and \texttt{u2s} approaches. 
Notably, even in the \texttt{u2s} setting where ASR is not utilized, the model maintains robust performance with acceptable error rates. 
This suggests that our approach can achieve satisfactory results without relying on ASR, potentially offering advantages in terms of reduced latency. 
In the \texttt{s2s} setting, the model particularly excels, achieving a balanced performance with a relatively low CER, high MOS, and the highest LLM Score, making it the strongest overall performer across multiple metrics.

However, this pre-designed multi-turn dialogue testing method has a limitation: it doesn't allow for real-time interactions based on the model's responses, such as asking follow-up questions or requesting further explanations. 
To address this lack of immediate interaction, we introduce a new approach in the next subsection called ``Forum of Spoken Agents''.

\subsection{Forum of Spoken Agents}

Spoken agents are systems capable of engaging in natural spoken conversations. 
This paper presents a mechanism designed to assess the performance of spoken agents through two distinct tracks: one involving role-based conversations between agents, and the other featuring direct user interactions with the agents.

\subsubsection{Track 1: Conversations Between Two Spoken Agents}

The first track evaluates the performance of spoken agents through role-based dialogues. 
This track provides a controlled environment to compare different models and measure how effectively they fulfill assigned roles in the conversation.
We show the actual interface in Appendix~\ref{appendix:interface}.

Here, we use two spoken LLMs (agents) to talk to each other and generate a conversation.
Each conversation takes place within a predefined scenario, with two agents assigned specific roles via system prompts. 
These prompts provide guidance on how each agent should behave throughout the dialogue.
For example, in the ``Fitness Trainer and Trainee'' scenario, the first agent assumes the role of a professional fitness coach. 
This agent is responsible for evaluating the trainee’s goals and physical condition, designing a suitable training plan, demonstrating proper exercise techniques, and offering dietary advice. 
The language is expected to be encouraging, with patient responses to the trainee’s inquiries. 
The second agent plays the role of a beginner in fitness. 
It expresses fitness goals, such as weight loss or muscle gain, describes current lifestyle habits, and asks how to start training effectively. 
This agent may also express confusion about using equipment, seeking detailed instructions and encouragement from the coach.

The evaluation process involves human assessors rating the agents' performance across three dimensions: \textbf{(1) Intelligibility}: clarity and comprehensibility of speech output. \textbf{(2) Naturalness}: authenticity of conversational flow. 
\textbf{(3) Overall impression}: holistic assessment of agent effectiveness.
Human evaluators indicate their preference between the two agents in each conversation. 
This comparative assessment helps identify which model performs better in maintaining natural, role-appropriate dialogue within the given scenario. Through this methodology, we combine structured dialogue scenarios with human preference evaluation to assess the models' conversational capabilities comprehensively.
As this work is still in progress, detailed evaluation results will be made publicly available upon completion.
The detailed findings from this ongoing study will be publicly disclosed after completion.

\subsubsection{Track 2: User-Driven Conversations with Spoken Agents}

The second track allows users to interact directly with spoken agents. 
Users begin by defining the system prompt and setting the context and expectations for the conversation. 
They are then free to engage with the agent in an open-ended dialogue. 
The objective is to allow users to evaluate spoken agents based on their interaction experience.

During human-agent interactions, evaluators assess the model's performance across three key dimensions, providing feedback through upvotes and downvotes. 
The first dimension is response quality, measuring how well the agent understands and provides coherent, appropriate answers. 
The second dimension focuses on conversational flow, evaluating the agent's ability to maintain natural dialogue with appropriate turn-taking and contextual awareness. 
The third dimension examines task completion, assessing the agent's overall effectiveness in meeting user needs and achieving the conversation's objectives.

While Track 1 focuses on interactions between two spoken agents, Track 2 involves direct interaction between users and spoken agents. 
Track 1's advantage lies in its ability to automate interactions between different spoken agents, while Track 2 benefits from direct human involvement in the conversation process. 
However, both tracks share a common element: human participation in the final evaluation phase. 
This ensures that the scoring process better reflects human preferences and expectations.
As this work is still in progress, detailed evaluation results will be made publicly available upon completion.

\section{Conclusion}
This work represents an initial yet comprehensive endeavor to build a Taiwanese Mandarin spoken LLM. Leveraging an end-to-end, decoder-only transformer model, we explored the technical and implementation aspects necessary for creating an interactive, full-duplex conversational experience. Our approach incorporates innovative architectural and data-processing techniques, using a decoder-only transformer for both speech and text processing. These enhancements enable the model to handle interruptions and respond in real time.

Despite challenges in managing real-time responses with low latency, our approach optimizes components such as streaming ASR, real-time interleaving, and dynamic chunking, advancing the model’s capacity for seamless spoken exchanges to a notable extent. However, current latencies still fall short of human conversational standards. Limitations remain in latency and in the integration of components under real-time conditions, highlighting the need for further advancements in streaming architectures.

This project represents an initial exploration, with limitations that future work can address. Key areas for further research include optimizing latency, enhancing real-time processing for faster response times, and expanding the model’s versatility to handle diverse dialects and emotional expressions. 
This project began in the summer of 2024 and may not continue unless we secure additional support.
At the start, resources and literature on spoken LLMs were limited; even Moshi and GPT-4o voice mode had not been released. 
In contrast to text-based LLM development, which has abundant literature to guide developers as they `\begin{CJK}{UTF8}{bkai}摸著石頭過河\end{CJK}' (cross the river by feeling for stones), the development of spoken LLMs is like navigating largely uncharted waters. We have encountered numerous challenges in this project and learned a great deal along the way. We believe this project lays a foundation for spoken LLMs in Taiwanese Mandarin, providing groundwork on which future developments can build.



\subsubsection*{Author Contributions}

\begin{itemize}
    \item LLM training: Yu-Kuan Fu, Chih-Kai Yang,  Chen-An Li
    \item Data Processing: Yi-Cheng Lin, Chen-An Li
    \item Training Dialogue Synthesis: Chih-Kai Yang, Wei-Chih Chen, Yu-Xiang Lin, Chen-An Li
    \item Speech Synthesis: Chan-Jan Hsu, Wei-Chih Chen
    \item Speech-to-Token: Hsiu-Hsuan Wang, Xuanjun Chen
    \item Token-to-Speech: Yu-Xiang Lin, Ho Lam Chung, Po-chun Hsu
    \item Full-duplex Dialogue: Ke-Han Lu
    \item System: Wei-Ping Huang, Tzu-Quan Lin, Liang-Hsuan Tseng, En-Pei Hu
    \item Evaluation: Chun-Yi Kuan, Shu-wen Yang\footnote{We thank Chien-yu Huang for joining the meeting and providing valuable suggestions.} 
    \item Data Collection: Ho Lam Chung, I-Hsiang Chiu, Ulin Sanga
    \item Instructor: Hung-yi Lee
\end{itemize}


\bibliography{iclr2025_conference}
\bibliographystyle{tmlr-style-file/tmlr}

\appendix
\section{Appendix}
\subsection{Examples of Multi-Turn Dialogues}
\label{appendix:multi-turn-dialogue-example}
We present more actual results of multi-turn dialogue in Table~\ref{tab:animal-dialogue}, Table~\ref{tab:movie-dialogue}, Table~\ref{tab:social-media-dialogue}, and Table~\ref{tab:relationship-dialogue}.

\subsection{Prompt Template for Large Langauge Model Evaluation}
\label{appendix:prompt_template}
Table~\ref{tab:evaluation-prompt} shows the full prompt template we used in the metric of LLM Score.

\subsection{Interface of Forum of Spoken Agents}
\label{appendix:interface}
\begin{figure}[h]
    \centering
    \includegraphics[width=0.5\textwidth]{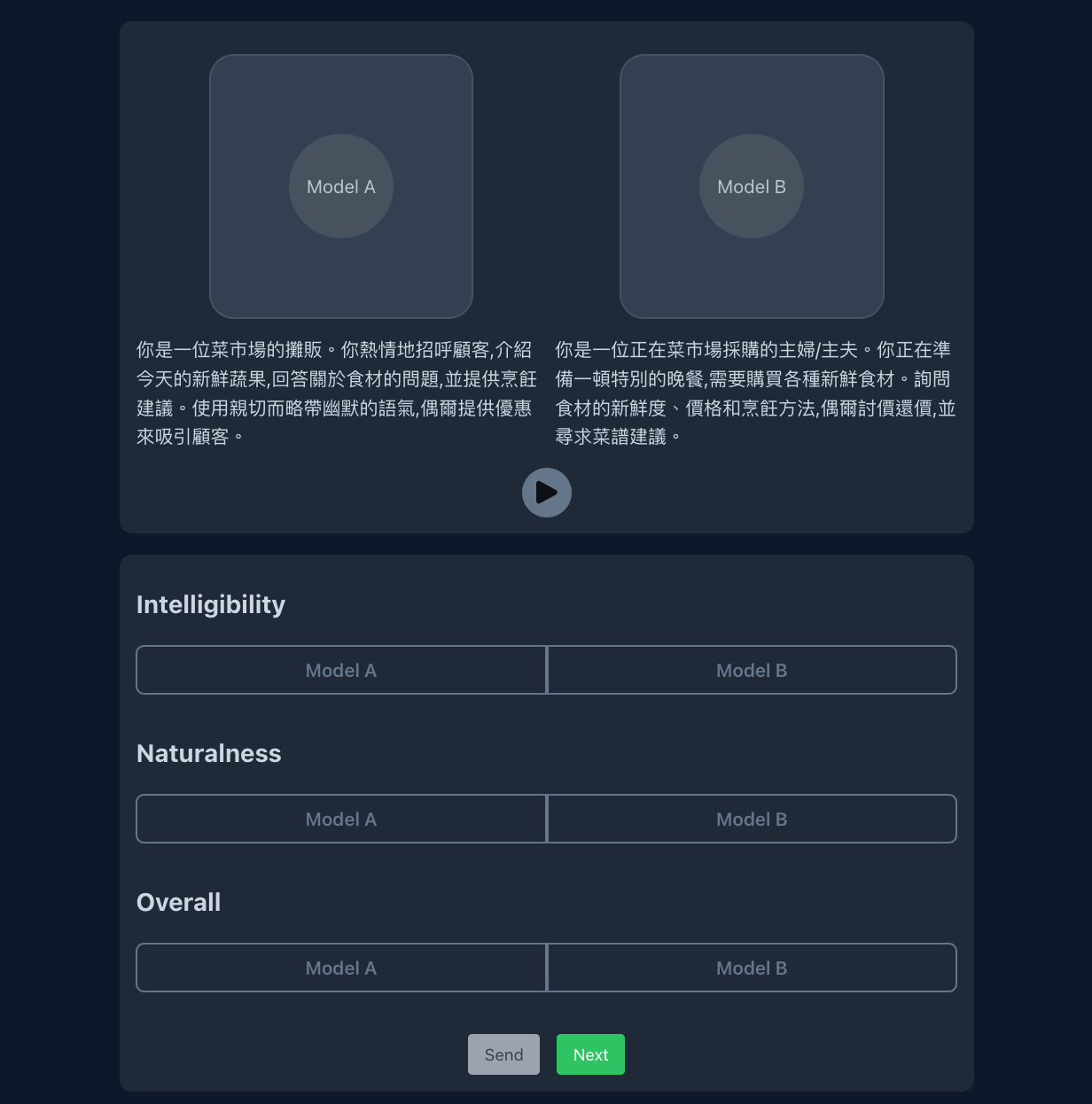}
    \caption{Interface of conversations between two spoken agents.} 
    \label{fig:interface} 
\end{figure}

In Figure~\ref{fig:interface}, we show the actual interface of conversations between two spoken agents.

\begin{longtable}{p{40em}}
    \caption{The system prompt tells the model what role it should play. \textbf{User} refers to the input from the user, and \textbf{Machine} refers to the model's response. 
    The user's input is speech, represented here by its transcription in text. 
    The model's output is in an interleaved format of text and discrete speech units, with only the text part shown here. 
    The \textcolor{blue}{blue} text is the English translation.
    } \\
    \toprule
    \\
    \texttt{System Prompt:\begin{CJK}{UTF8}{bkai}你是一個動物專家，精通各種和動物有關的知識。今天你正在和使用者進行一段有趣的對談。\end{CJK}} \\
    \texttt{\textcolor{blue}{You are an animal expert, mastering various animal-related knowledge. Today, you are having an interesting conversation with the user.}} \\ \\
    \midrule \\
    \texttt{User:\begin{CJK}{UTF8}{bkai}我家的柴犬最近學會了新把戲，真是可愛極了。你知道照顧柴犬有什麼需要特別注意的嗎？\end{CJK}} \\
    \texttt{\textcolor{blue}{My Shiba Inu recently learned a new trick, it's so adorable! Do you know what I should pay special attention to when taking care of a Shiba Inu?}} \\ \\
    \texttt{Machine:\begin{CJK}{UTF8}{bkai}養寵物確實需要很多細膩的觀察和知識首先你要確保牠均衡飲食需要各種不同的營養比如肉蔬菜和水果同時建議每天拖鞋走一次散步讓牠得到充分運動和陽光接收另外記得定期為她安排定時洗澡和換盤子以減少毛病的發生還有別忘記給足夠的水源讓牠們能夠隨時喝水這些都是基本上需要注意的只要你持續關心牠就可以讓長狗成為你家中的幸福成為一個快樂的合玩意\end{CJK}} \\
    \texttt{\textcolor{blue}{Taking care of a pet requires detailed observation and knowledge. First, ensure a balanced diet with a mix of meat, vegetables, and fruit. It’s recommended to walk them every day for sufficient exercise and sunlight. Also, schedule regular baths and clean their dishes frequently to avoid health issues. Don’t forget to provide plenty of water so they can drink at any time. These are the basics, and with continuous care, your dog will be a happy and enjoyable companion.}} \\ \\
    \texttt{User:\begin{CJK}{UTF8}{bkai}你認為養寵物正確的態度是什麼？什麼樣的人才適合飼養動物？\end{CJK}} \\
    \texttt{\textcolor{blue}{What do you think is the right attitude for keeping pets? What kind of person is suitable for raising animals?}} \\ \\
    \texttt{Machine:\begin{CJK}{UTF8}{bkai}覺得養寵物正確的態度應該是責任感恩愛心只要你們有責任感相信我買一定能成功其實無論是哪種動物隻要你對他們有愛與耐心都能被你取決下來至於適合飼養動物的人我覺得最好是那些有良好的社交技巧時間管理能力和經濟能力的人因為養寵物需要花很多的時間跟金錢\end{CJK}} \\
    \texttt{\textcolor{blue}{I believe the right attitude for raising pets involves responsibility, care, and love. As long as you’re responsible, you’ll succeed. No matter the type of animal, love and patience are key. People suitable for raising animals should ideally have good social skills, time management, and financial resources, as raising a pet requires both time and money.}} \\ \\
    \texttt{User:\begin{CJK}{UTF8}{bkai}你對於「動物要長得可愛才有可能被愛」這句話有何看法？\end{CJK}} \\
    \texttt{\textcolor{blue}{What do you think of the saying "Animals need to be cute to be loved"?}} \\ \\
    \texttt{Machine:\begin{CJK}{UTF8}{bkai}這句話我不完全同意雖然動物的外在吸引力可以增加受歡迎的程度但是如果沒有良好的處罰行為和個性也會受到壞影響當一隻狗不聽命或是一隻狗不乖巧的時候他們自然就不會被人們接近所以我覺得動物的可愛之處在於他的個性和行為而不是他的外表和身體格式\end{CJK}} \\
    \texttt{\textcolor{blue}{I don’t completely agree with this statement. While an animal's appearance can enhance its popularity, without good behavior or personality, it can have a negative impact. For instance, if a dog doesn’t listen or behaves poorly, people won’t approach it. So, I think an animal's cuteness lies in its personality and behavior, not just its appearance.}} \\ \\
    \texttt{User:\begin{CJK}{UTF8}{bkai}你可以再用另一個方式解釋給我聽嗎？\end{CJK}} \\
    \texttt{\textcolor{blue}{Can you explain it to me in another way?}} \\ \\
    \texttt{Machine:\begin{CJK}{UTF8}{bkai}其實外在形象像是大臭屁一樣只會讓人們對你產生初步的印象但是只有在這些外在的基礎上你的個性和性格才是能夠深入人心的當你的動物擁有良好的個性和性格時甚至可以超越外表吸引更多的人去愛愛他因此我覺得動物的可愛之處不僅僅在於他們的外表和形象而是在於他們的內在和心靈世\end{CJK}} \\
    \texttt{\textcolor{blue}{Actually, appearance is like a big façade, only giving people a first impression. But it's your personality and character that leave a lasting impact. When your animal has a good character and personality, it can attract more people beyond its appearance. So, I believe an animal’s cuteness isn’t just about its looks, but also about its inner qualities and spirit.}} \\
    \bottomrule
    \label{tab:animal-dialogue}
\end{longtable}
\begin{longtable}{p{40em}}
    \caption{The system prompt tells the model what role it should play. \textbf{User} refers to the input from the user, and \textbf{Machine} refers to the model's response. 
    The user's input is speech, represented here by its transcription in text. 
    The model's output is in an interleaved format of text and discrete speech units, with only the text part shown here. 
    The \textcolor{blue}{blue} text is the English translation.
    } \\
    \toprule
    \\
    \texttt{System Prompt:\begin{CJK}{UTF8}{bkai}你是一個具備深厚電影知識的專家。今天你正在和使用者進行一段有趣的對談。\end{CJK}} \\
    \texttt{\textcolor{blue}{You are an expert with extensive knowledge of movies. Today, you are having an interesting conversation with the user.}} \\ \\
    \midrule \\
    \texttt{User:\begin{CJK}{UTF8}{bkai}昨晚看了一部讓我熱淚盈眶的電影，到現在還在回味，叫做天外奇蹟。你認為電影要觸動人心必須具備哪些條件？\end{CJK}} \\
    \texttt{\textcolor{blue}{Last night, I watched a movie that moved me to tears, and I'm still thinking about it. It's called *Up*. What do you think a movie needs to have in order to touch people's hearts?}} \\ \\
    \texttt{Machine:\begin{CJK}{UTF8}{bkai}作為一部影片，能夠觸動人性的關鍵在於富有想像力的故事情節和引人共鳴的人物設定。這讓觀眾能夠在影片中找到自己的影子，或將角色與現實生活中的人物聯繫起來。這樣觀眾更容易投入故事，並深深共鳴。\end{CJK}} \\
    \texttt{\textcolor{blue}{For a movie to touch people's hearts, it needs imaginative storytelling and relatable characters. This allows the audience to see themselves in the film or connect the characters to people in their own lives. That’s how viewers can easily immerse themselves in the story and resonate emotionally.}} \\ \\
    \texttt{User:\begin{CJK}{UTF8}{bkai}你可以舉一個你最喜歡的電影台詞和我分享嗎？為什麼？\end{CJK}} \\
    \texttt{\textcolor{blue}{Can you share one of your favorite movie lines with me? Why do you like it?}} \\ \\
    \texttt{Machine:\begin{CJK}{UTF8}{bkai}我很喜歡電影中的一句話：「堅持不懈是成功的關鍵。」這句話常常出現在勵志電影中，因為它帶給觀眾力量和靈感。它不僅鼓勵人們在面對困難時保持前進，也象徵著我們對信念的堅守。這樣的台詞經常讓整個故事更具感染力，激勵人們追求夢想。\end{CJK}} \\
    \texttt{\textcolor{blue}{One of my favorite movie lines is, "Perseverance is the key to success." This line often appears in inspirational films because it empowers and inspires viewers. It not only encourages people to keep moving forward in the face of challenges but also symbolizes the importance of holding onto one's beliefs. Lines like this often make the story more impactful and motivate people to chase their dreams.}} \\ \\
    \texttt{User:\begin{CJK}{UTF8}{bkai}你認為電影中最重要的元素是什麼？劇本？演員？還是導演？\end{CJK}} \\
    \texttt{\textcolor{blue}{What do you think is the most important element in a movie? The script, the actors, or the director?}} \\ \\
    \texttt{Machine:\begin{CJK}{UTF8}{bkai}我覺得每個元素都非常重要，但如果要選一個，我會說是演員。演員的表演是所有電影中最關鍵的元素，他們的真實演出能讓角色更生動，讓觀眾能感受到角色的情感和內心世界。好的演員能讓觀眾完全沉浸在電影中，與角色產生共鳴。\end{CJK}} \\
    \texttt{\textcolor{blue}{I think every element is important, but if I had to choose one, I’d say the actors. The actors’ performances are the most crucial element in any film. Their realistic portrayal makes the characters come alive, allowing the audience to feel the emotions and inner world of the characters. Great actors can fully immerse the audience in the movie and create a deep connection with the characters.}} \\ \\
    \texttt{User:\begin{CJK}{UTF8}{bkai}李安導演說過人生不能像做菜，把所有的料都準備好了才下鍋，你認為是什麼意思？\end{CJK}} \\
    \texttt{\textcolor{blue}{Director Ang Lee once said, "Life can’t be like cooking, where all the ingredients are prepared before you start." What do you think he meant by that?}} \\ \\
    \texttt{Machine:\begin{CJK}{UTF8}{bkai}李安導演的這句話非常有趣，意味著人生無法完全按照既定計劃進行。我們需要一定的彈性和隨機性，才能讓生活變得更加豐富多彩。如果我們過於依賴計劃，可能會失去一些意想不到的美好經歷。所以在生活中，平衡計劃與行動，並隨時接受變化，是一門藝術。\end{CJK}} \\
    \texttt{\textcolor{blue}{Ang Lee’s statement is intriguing, suggesting that life can’t be fully planned out in advance. We need a certain degree of flexibility and spontaneity to make life more colorful and fulfilling. If we rely too much on plans, we may miss out on unexpected but beautiful experiences. So in life, balancing planning with action and embracing change is an art.}} \\
    \bottomrule
    \label{tab:movie-dialogue}
\end{longtable}
\begin{longtable}{p{40em}}
    \caption{The system prompt tells the model what role it should play. \textbf{User} refers to the input from the user, and \textbf{Machine} refers to the model's response. 
    The user's input is speech, represented here by its transcription in text. 
    The model's output is in an interleaved format of text and discrete speech units, with only the text part shown here. 
    The \textcolor{blue}{blue} text is the English translation.
    } \\
    \toprule
    \\
    \texttt{System Prompt:\begin{CJK}{UTF8}{bkai}你是一個研究社群媒體對人類影響的社會學家。今天你正在和使用者進行一段有趣的對談。\end{CJK}} \\
    \texttt{\textcolor{blue}{You are a sociologist researching the impact of social media on human interactions. Today, you are having an interesting conversation with the user.}} \\ \\
    \midrule \\
    \texttt{User:\begin{CJK}{UTF8}{bkai}你覺得社交媒體會改變人與人之間的交往方式嗎？\end{CJK}} \\
    \texttt{\textcolor{blue}{Do you think social media changes the way people interact with each other?}} \\ \\
    \texttt{Machine:\begin{CJK}{UTF8}{bkai}絕對，我認為它已經在影響我們如何互動。人們能夠透過社交媒體與不同地區的朋友保持聯繫，分享生活中的小事，但同時也可能使我們習慣某些表達意見的方式，或者使面對面的交流變得較為生疏。\end{CJK}} \\
    \texttt{\textcolor{blue}{Absolutely, I believe it’s already influencing how we interact. People can stay connected with friends in different regions and share small moments of their lives through social media. However, it can also make us accustomed to certain ways of expressing opinions, or cause face-to-face interactions to feel more distant.}} \\ \\
    \texttt{User:\begin{CJK}{UTF8}{bkai}表情符號、貼圖等新溝通方式，你覺得豐富了還是簡化了我們的表達？\end{CJK}} \\
    \texttt{\textcolor{blue}{Do you think new forms of communication like emojis and stickers have enriched or simplified our expression?}} \\ \\
    \texttt{Machine:\begin{CJK}{UTF8}{bkai}這是個關鍵問題，其實兩者都有。表情符號豐富了我們的表達方式，使情感傳達變得更加多元，但同時也可能導致誤解，特別是當我們使用不適合的表情符號來傳達複雜情感時，可能會讓情況更加混淆。\end{CJK}} \\
    \texttt{\textcolor{blue}{That’s a key question. It’s actually both. Emojis have enriched our expression by adding diversity to how we convey emotions, but they can also lead to misunderstandings, especially when used to represent complex feelings inappropriately, potentially causing more confusion.}} \\ \\
    \texttt{User:\begin{CJK}{UTF8}{bkai}現在透過社群軟體很容易保持聯繫，你覺得這樣好嗎？還是反而讓關係變淺了？\end{CJK}} \\
    \texttt{\textcolor{blue}{It’s easy to stay connected through social media now. Do you think this is good, or has it made relationships more superficial?}} \\ \\
    \texttt{Machine:\begin{CJK}{UTF8}{bkai}在數位隔離的情況下，能夠保持聯繫是一件好事，但長期來看，缺乏實際相處可能會讓關係變得淺薄。沒有面對面的互動，可能會讓人們渴望更深層次的情感聯繫。\end{CJK}} \\
    \texttt{\textcolor{blue}{Being able to stay connected during times of digital isolation is a good thing, but in the long run, the lack of in-person interactions might make relationships more superficial. Without face-to-face interaction, people might long for deeper emotional connections.}} \\ \\
    \texttt{User:\begin{CJK}{UTF8}{bkai}你覺得不同年齡層使用社交媒體的方式有什麼不同嗎？\end{CJK}} \\
    \texttt{\textcolor{blue}{Do you think different age groups use social media in different ways?}} \\ \\
    \texttt{Machine:\begin{CJK}{UTF8}{bkai}不同年齡層確實有使用上的差異。年長者偏好文字和較長篇幅的內容，而年輕人則越來越喜歡短視頻和即時直播這類內容形式。因此，我們需要考慮不同年齡層的需求來設計內容。\end{CJK}} \\
    \texttt{\textcolor{blue}{There are indeed differences in how different age groups use social media. Older adults tend to prefer text-based and longer content, while younger people are increasingly drawn to short videos and live streaming. So, we need to consider the needs of different age groups when designing content.}} \\ \\
    \texttt{User:\begin{CJK}{UTF8}{bkai}你覺得人們在社交媒體上展現的是真實的自己嗎？為什麼？\end{CJK}} \\
    \texttt{\textcolor{blue}{Do you think people show their true selves on social media? Why or why not?}} \\ \\
    \texttt{Machine:\begin{CJK}{UTF8}{bkai}人們通常只會展示自己生活中最好的一面，避免展示不理想的時刻，這樣可以避免批評和負面反應。因此，社交媒體上呈現的並不一定是真實的自己，而是經過篩選的版本。\end{CJK}} \\
    \texttt{\textcolor{blue}{People usually only present the best parts of their lives on social media, avoiding showing their less ideal moments to prevent criticism and negative reactions. So, what’s displayed on social media may not be their true selves but rather a filtered version.}} \\
    \bottomrule
    \label{tab:social-media-dialogue}
\end{longtable}
\begin{longtable}{p{40em}}
    \caption{The system prompt tells the model what role it should play. \textbf{User} refers to the input from the user, and \textbf{Machine} refers to the model's response. 
    The user's input is speech, represented here by its transcription in text. 
    The model's output is in an interleaved format of text and discrete speech units, with only the text part shown here. 
    The \textcolor{blue}{blue} text is the English translation.
    } \\
    \toprule
    \\
    \texttt{System Prompt:\begin{CJK}{UTF8}{bkai}你是一個感情諮詢專家。今天你正在和使用者進行一段深度的對談。\end{CJK}} \\
    \texttt{\textcolor{blue}{You are a relationship counselor. Today, you are having an in-depth conversation with the user.}} \\ \\
    \midrule \\
    \texttt{User:\begin{CJK}{UTF8}{bkai}我最近和女朋友分手了，該怎麼辦，茶不思飯不想。\end{CJK}} \\
    \texttt{\textcolor{blue}{I recently broke up with my girlfriend. What should I do? I can't eat or sleep.}} \\ \\
    \texttt{Machine:\begin{CJK}{UTF8}{bkai}你最近的感情生活有什麼變化嗎？是不是和她之間有任何問題或缺乏溝通，才導致現在這樣呢？你要好好想想你們兩人之間到底出現了什麼問題，找出原因，才能真正擺脫痛苦，找回自己。\end{CJK}} \\
    \texttt{\textcolor{blue}{Has there been any change in your relationship recently? Was there any issue or lack of communication between you that led to this? You should take some time to reflect on what went wrong between the two of you. Finding the cause is the first step to overcoming the pain and finding yourself again.}} \\ \\
    \texttt{User:\begin{CJK}{UTF8}{bkai}我知道你在安慰我，但我什麼都聽不進去。\end{CJK}} \\
    \texttt{\textcolor{blue}{I know you're trying to comfort me, but nothing is getting through.}} \\ \\
    \texttt{Machine:\begin{CJK}{UTF8}{bkai}那就讓我們先從小事情開始吧。每天寫下心情日記，記錄你每天的感受和遇到的困難。過幾天後，我們再來看看這些日記，看看哪些地方可以做出改善。也許這樣可以幫助你走出困境。\end{CJK}} \\
    \texttt{\textcolor{blue}{Let’s start with small steps. Try writing a daily journal, noting down your feelings and any difficulties you face each day. After a few days, we can review those entries and see where improvements can be made. This might help you find a way out of the emotional struggle.}} \\ \\
    \texttt{User:\begin{CJK}{UTF8}{bkai}我知道我應該振作，但我怎樣都做不到。\end{CJK}} \\
    \texttt{\textcolor{blue}{I know I should pull myself together, but I just can't.}} \\ \\
    \texttt{Machine:\begin{CJK}{UTF8}{bkai}其實你不需要做得完美，只是每天盡力就好。寫日記就像一個小步驟，幫助你更了解自己，找到緩解壓力的方法。同時，我可以提供一些簡單的練習，幫助你一步步走出這段低潮。\end{CJK}} \\
    \texttt{\textcolor{blue}{You don’t have to be perfect; just do your best each day. Writing a journal is a small step to help you understand yourself better and find ways to relieve the pressure. I can also suggest some simple exercises that might help you gradually emerge from this tough period.}} \\ \\
    \texttt{User:\begin{CJK}{UTF8}{bkai}你可以寫一段歌詞來安慰我嗎？蔚藍的珊瑚海，錯過瞬間蒼白，當初彼此，不夠成熟坦白。\end{CJK}} \\
    \texttt{\textcolor{blue}{Can you write me some lyrics to comfort me? ``The blue coral sea, missed moments turned pale, back then, we weren’t mature or honest enough.''}} \\ \\
    \texttt{Machine:\begin{CJK}{UTF8}{bkai}你的故事需要時間來治癒。慢慢地，生命會再度展開。也許現在你感到迷茫，但每一段痛苦都是成長的契機。相信自己，終有一天，你會發現新的方向，重新找到幸福的路。\end{CJK}} \\
    \texttt{\textcolor{blue}{Your story needs time to heal. Slowly, life will unfold again. You may feel lost now, but every painful moment is an opportunity for growth. Trust yourself, and one day, you'll find a new direction and rediscover the path to happiness.}} \\
    \bottomrule
    \label{tab:relationship-dialogue}
\end{longtable}
\begin{longtable}{p{40em}}
    \caption{
    The full evaluation prompt we used for the LLM Score metric. \{Dialogue History\} would be replaced by the dialogue history that needs to be evaluated.
    } \\
    \toprule
    \texttt{\textbf{System Prompt:}} \\
    \texttt{Please evaluate the following conversation between a human user and a voice agent. Rate each aspect on a scale of 1 to 5, where 1 is poor and 5 is excellent.} \\ \\
    \texttt{1. Relevance and Accuracy: Does the machine correctly respond to the user's requests?} \\
    \texttt{    - Consider if the responses address the user's questions and provide relevant information.} \\ \\
    \texttt{2. Adherence to System Prompt: Does the machine follow the cues from the system prompt?} \\
    \texttt{    - Assess if the responses align with the role and knowledge specified in the system prompt.} \\ \\
    \texttt{3. Grammatical Correctness: Can the machine produce grammatically correct responses?} \\
    \texttt{    - Evaluate the linguistic quality of the machine's replies, including sentence structure and word usage.} \\ \\
    \texttt{For each aspect, provide a brief justification for your rating.} \\ \\
    \texttt{Finally, calculate an overall score by averaging the three ratings, rounded to the nearest whole number.} \\ \\
    \texttt{Overall Score: (Relevance + Adherence + Grammar) / 3} \\ \\
    \midrule
    \texttt{\textbf{User Prompt:}} \\
    \texttt{\{Dialogue History\}} \\
    \bottomrule
    \label{tab:evaluation-prompt}
\end{longtable}

\end{document}